\documentclass[%
 reprint,
 amsmath,amssymb,
 aps,
]{revtex4-2}

\usepackage{graphicx}%
\usepackage{dcolumn}%
\usepackage{bm}%

\usepackage{color}
\usepackage{glossaries}
\usepackage{xspace}

\usepackage{pdfpages}
\usepackage{pgffor}
\makeatletter
\AtBeginDocument{\let\LS@rot\@undefined}
\makeatother

\newacronym{gmc}{GMC}{Galilean Monte Carlo}
\newacronym{sd}{SD}{Sinkhorn divergence}
\newacronym{rhmc}{RHMC}{Reflective Hamiltonian Monte Carlo with inexact reflections}

\newcommand{\dimension}{\ensuremath{n}\xspace}
\newcommand{\timestep}{\ensuremath{\tau{}}\xspace}
\newcommand{\unittime}{\ensuremath{\mathrm{MCS}}\xspace}

\DeclareMathOperator{\sgn}{sgn}

\begin{document}

\preprint{APS/123-QED}

\title{Resonances in reflective Hamiltonian Monte Carlo}

\author{Namu Kroupa$^{1,2,3}$}
\email{nk544@cam.ac.uk}
\author{G\'abor Cs\'anyi$^{3}$}%
\author{Will Handley$^{2,4}$}
\affiliation{$^1$Astrophysics Group, Cavendish Laboratory, J.J. Thomson Avenue, Cambridge, CB3 0HE, UK}
\affiliation{$^2$Kavli Institute for Cosmology, Madingley Road, Cambridge, CB3 0HA, UK} 
\affiliation{$^3$Engineering Laboratory, University of Cambridge, Cambridge CB2 1PZ, UK}
\affiliation{$^4$Institute of Astronomy, University of Cambridge, Madingley Road CB3 0HA, UK}

\date{\today}

\begin{abstract}

In high dimensions, reflective Hamiltonian Monte Carlo with inexact reflections exhibits slow mixing when the particle ensemble is initialised from a Dirac delta distribution and the uniform distribution is targeted. By quantifying the instantaneous non-uniformity of the distribution with the Sinkhorn divergence, we elucidate the principal mechanisms underlying the mixing problems. In spheres and cubes, we show that the collective motion transitions between fluid-like and discretisation-dominated behaviour, with the critical step size scaling as a power law in the dimension. In both regimes, the particles can spontaneously unmix, leading to resonances in the particle density and the aforementioned problems.
Additionally, low-dimensional toy models of the dynamics are constructed which reproduce the dominant features of the high-dimensional problem.
Finally, the dynamics is contrasted with the exact Hamiltonian particle flow
and tuning practices are discussed.

\end{abstract}

\maketitle

\section{\label{sec:level1}Introduction}

\gls{rhmc}
is an algorithm used to sample from uniform distributions in $\mathbb{R}^\dimension$. Originally introduced to perform slice sampling~\cite{neal2003slice}, it was later adopted within the nested sampling algorithm~\cite{skilling2006nested} to calculate the normalising constant of a probability distribution, for which it is used in the physical sciences, such as
in Bayesian inference and materials science~\cite{feroz2009multinest, partay2010efficient, partay2021nested, ashton2022nested, buchner2023nested}.
Inexact reflections are necessary
when the boundary of the uniform distribution is not known \textit{a priori} and solving numerically for the intersection of a particle trajectory with the boundary to perform exact reflections is computationally prohibitive. Several variations of the algorithm have been proposed~\cite{neal2003slice, skilling2012bayesian, skilling2019galilean}, differing in the way the reflection is performed and the time at which the momentum of a particle is randomised.

However, in the context of nested sampling, it is observed that \gls{rhmc} introduces a negative systematic error in the normalising constant which increases in magnitude with the dimensionality of the distribution~\cite{lemos2023improving, olander2020constrained}.
The algorithm therefore does not seem to scale beyond $\mathcal{O}(10)$ dimensions in practice,
which has been attributed to poor mixing~\cite{lemos2023improving} and the curse of dimensionality~\cite{olander2020constrained}.
It has been observed that the addition of Gaussian noise to the particle momentum leads to a decrease in the error~\cite{lemos2023improving}, thus rendering the dynamics diffusive
and lowering the degree of coherence in a Markov chain. However, it has remained elusive why coherence induces the observed problems in the first place. 
While existing algorithms not using gradient information, notably hit-and-run slice sampling~\cite{smith1984efficient}, do not exhibit such an error~\cite{handley2015polychord}, the overall increase in availability of gradient information due to automatic differentiation frameworks~\cite{baydin2018automatic, paszke2019pytorch} paired with an inflation in parameter space dimensionality~\cite{ruiz2023analytical} has increased interest in algorithms which utilise gradients in an effective manner. This is particularly the case within computational materials science, wherein forces are already necessary for molecular dynamics simulations~\cite{unke2021machine} and reaching the thermodynamic limit is a primary aim.
While the suboptimal performance of \gls{rhmc} has prompted changes in how the reflective update is conducted~\cite{skilling2019galilean}, the precise cause has remained unidentified and consequently unaddressed to date.

The problem is exacerbated by the fact that, in contrast to the usual setting of Markov Chain Monte Carlo, in which a chain is run until the burn-in time is reached and correlations between subsequent samples are removed by appropriate subsampling~\cite{brooks2011handbook}, the application in nested sampling falls into the many-short-chains regime, in which multiple chains are run in parallel until a single sufficiently decorrelated sample per chain is obtained.
More specifically, the chains are initialised from a mixture of Dirac distributions centred at a prepared set of points inside a volume, called live points,
from which current state-of-the-art implementations initialise chains in parallel. Typically the number of live points is set proportional to $\dimension$ and number of chains is chosen to be greater than or equal to the number of live points.
To isolate the mixing problems, we initialise a large number of chains from a single point, corresponding to a setting with a single live point and many parallel processors. A setting with multiple live points thus amounts to an equally weighted superposition of the particle densities observed in our setting.
However, even when the number of live points scales linearly with $\dimension$, an exponential proliferation of minima in realistic energy landscapes~\cite{stillinger1999exponential, hoare1976statistical, tsai1993use, pickard2011ab} causes the live point population to drop below a single live point per basin on average. Since basins become ergodically separated during a nested sampling run, our setting of a single isolated live point becomes typical.
Hence, even in a practical setting, the question of convergence therefore reduces to the short time-scale mixing of an ensemble of particles.

In this respect, guarantees for asymptotic convergence, as provided by ergodic theorems, are uninformative. 
Instead, we require tight bounds on the mixing time of~\gls{rhmc}, which do not exist for the setting under consideration.
Typically, an upper bound on the mixing time is dependent on the warmness of the initial distribution~\cite{lovasz2004hit}, which quantifies a notion of the difference from the stationary distribution. For a Dirac delta initialisation, the warmness is infinite.
Application of the results in~\cite{lovasz1993random, kannan2006blocking, lovasz1990mixing, vempala2005geometric}, which provide an upper bound in terms of the warmness, therefore give a vacuous upper bound on the mixing time.
Instead, the particle distribution after one step of the algorithm may be used, which renders the warmness finite. However, the bounds also depend on the conductance of the Markov Chain, for which there are currently no results available to the authors' knowledge.

We therefore measure the mixing of the Markov Chain computationally. For this, we use the \gls{sd} to quantify the distance of the density of an ensemble of particles to the uniform distribution and track the \gls{sd} as a function of time. 
We focus on an instance of \gls{rhmc} known as \gls{gmc}~\cite{skilling2012bayesian} as it is most prevalently used in practice~\cite{feroz2013exploring, partay2021nested}. The alternative variant known as Reflective Slice Sampling~\cite{neal2003slice}, which re-randomises the momentum upon rejection of a step, follows the same dynamics as \gls{gmc} in the sphere as the dynamics is rejection-free.
We investigate mixing in the sphere and cube in $\dimension$ dimensions, which pose tractable cases due to their high symmetry.
We study the dynamics in the sphere as it is relevant to the case of isotropic probability distributions in nested sampling and slice sampling. 
Moreover, linear statistical models with Gaussian errors on the data have Gaussian likelihoods in parameter space. For non-linear models, the Laplace approximation around the maximum likelihood estimator yields a Gaussian.
In the context of Boltzmann sampling in computational materials science, a harmonic approximation of a minimum in the potential energy surface leads to the Gaussian case as well. Since current state-of-the-art nested sampling codes map non-isotropic Gaussian distributions to an isotropic one via a coordinate transformation~\cite{handley2015polychord}, the sphere is an archetypal case to be investigated.
The investigation is extended to the cube as its faces are flat. In applications, flat boundaries occur in polytope volume calculations~\cite{emiris2014efficient, chevallier2022efficient}, Hamiltonian Monte Carlo on discontinuous distributions~\cite{mohasel2015reflection}, when nested sampling is used for distributions with discrete parameters~\cite{hee2016bayesian, kroupa2024kernel} and in toy problem likelihoods with plateaus~\cite{fowlie2021nested}.

The paper is structured as follows. Section~\ref{sec:background} introduces \gls{gmc} and \gls{sd}. In Sections~\ref{sec:dynamics-in-unit-sphere} and~\ref{sec:dynamics-in-cube}, the dynamics of \gls{gmc} in the sphere and cube is analysed, respectively, and discussed in Section~\ref{sec:discussion}. The paper concludes with Section~\ref{sec:conclusions}. Appendices~\ref{sec:appendix-explicit-construction}, \ref{sec:appendix-derivation-circle-density} and~\ref{sec:appendix-derivation-of-the-supersonic-frequency} provide derivations, Appendix~\ref{sec:appendix-power-spectral-density-sphere-larger-range} shows additional results for the sphere and the scaling of the mean chord length with dimension is calculated in Appendix~\ref{sec:appendix-scaling-of-mean-chord-length}.

\section{Background}\label{sec:background}

In this section, we introduce \gls{gmc} and Sinkhorn divergences. We will subsequently use Sinkhorn divergences to measure the convergence of the \gls{gmc} algorithm. 

\subsection{Galilean Monte Carlo}\label{sec:gmc}

\begin{figure}
	\centering
	\includegraphics[width=\columnwidth]{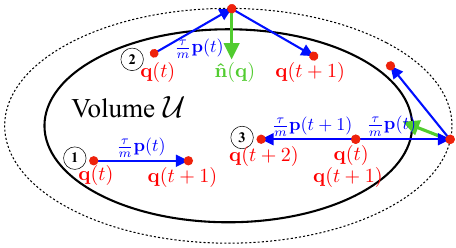}
	\caption{Possible moves of the GMC Markov Chain. A particle starting at $\mathbf{q}(t)$ with momentum $\mathbf{p}(t)$ inside a volume $\mathcal{U}$ attempts to move forward (1), reflect (2) or reverse its direction (3). Since the motion is discretised, the particle will overstep the boundary of $\mathcal{U}$ so that the normal vector $\mathbf{\hat{n}}(\mathbf{q})$ on the outside is used for reflections. This necessitates the definition of $\mathbf{\hat{n}}(\mathbf{q})$ as a vector field outside $\mathcal{U}$, which we define by considering scaled copies of the boundary (dotted line).}
	\label{fig:gmc-figure}
\end{figure}

Starting from an initial position and momentum inside the volume $\mathcal{U}$, \gls{gmc}~\cite{skilling2012bayesian} either moves forward, reflects using the normal vector $\mathbf{\hat{n}}$ or reverses the initial momentum (Figure~\ref{fig:gmc-figure}). It tries these moves in this order and only tries the next move if the proposed point lies outside the volume. 
This defines a step in the \gls{gmc} algorithm.
More precisely, given an initial position $\mathbf{q}(t)$, a momentum $\mathbf{p}(t)$, a time step \timestep and a scalar mass $m$, the position and momentum at time $t+1$ are calculated as follows:
\begin{enumerate}
	\item Let $\mathbf{q}_1=\mathbf{q}(t)+\frac{\tau}{m}\mathbf{p}(t)$. If $\mathbf{q}_1$ is inside, proceed with $\mathbf{q}(t+1)=\mathbf{q}_1$ and $\mathbf{p}(t+1)=\mathbf{p}(t)$.
	\item Otherwise, let ${\mathbf{\hat{n}}_1=\mathbf{\hat{n}}(\mathbf{q}_1)}$, ${\mathbf{p}_1=\mathbf{p}(t)-2(\mathbf{p}(t)^\top\mathbf{\hat{n}}_1)\mathbf{\hat{n}}_1}$ and $\mathbf{q}_2=\mathbf{q}_1+\frac{\tau}{m}\mathbf{p}_1$. If $\mathbf{q}_2$ is inside, proceed with $\mathbf{q}(t+1)=\mathbf{q}_2$ and $\mathbf{p}(t+1)=\mathbf{p}_1$.
	\item Otherwise, proceed with ${\mathbf{q}(t+1)=\mathbf{q}(t)}$ and ${\mathbf{p}(t+1)=-\mathbf{p}(t)}$.
\end{enumerate}
We call the second and third branches the reflection and rejection branches, respectively.
Note that \gls{gmc} reflects inexactly as it uses a normal vector strictly outside the volume, as opposed to dynamical billiards which uses the normal vector at the boundary. 

We define a trajectory as a sequence of $L$ steps, mapping the particle state from time $t$ to $t+L$. The \gls{gmc} algorithm proceeds by
sampling a momentum from the Gaussian distribution $\mathcal{N}(\mathbf{0},\sigma_p^2\mathbf{1})$, where $\mathbf{0}$ and $\mathbf{1}$ are the zero vector and the identity matrix of appropriate sizes, respectively, and $\sigma_p$ is the momentum standard deviation, followed by a trajectory and repeating this by sampling a new momentum, i.e. re-randomising the momentum.
By the term \gls{gmc} dynamics, we refer to the deterministic evolution of a particle in a trajectory.
In \cite{supp} (see also \cite{brofos2021numerical, brooks2011handbook, brofos2021evaluating, skilling2019galilean, neal2003slice} therein), we prove that any 
variety of \gls{rhmc}
has the uniform distribution as its stationary distribution and show that this also holds for \gls{gmc} as a special case.
Finally, we note that momentum re-randomisations are required for ergodicity although we do not prove this here.

\gls{gmc} requires us to define a vector field $\mathbf{\hat{n}}(\mathbf{q})$ which matches the boundary normal vector at the boundary. While many choices are possible, for convex boundaries we can imagine moving each point on the boundary at constant speed along its instantaneous normal vector. This constructs a family of boundaries, indexed by time. The union of the normal vector fields of such boundaries defines our normal vector field $\mathbf{\hat{n}}(\mathbf{q})$ (Figure~\ref{fig:gmc-figure}).
For the sphere and cube, we give the explicit form thereof in Sections~\ref{sec:dynamics-in-unit-sphere} and~\ref{sec:dynamics-in-cube}, respectively.
In the application within the nested sampling algorithm, $\mathbf{\hat{n}}(\mathbf{q})$ is naturally defined by the gradient of the likelihood~\cite{skilling2006nested}, $\nabla \mathcal{L}(\mathbf{q})$, and the volume $\mathcal{U}$ corresponds to a level set of the likelihood function, $\{\mathbf{q}\mid \mathcal{L}(\mathbf{q})<\mathcal{L}_\star\}$, for some fixed value of the likelihood $\mathcal{L}_\star$. In the application within slice sampling~\cite{neal2003slice}, the volume and $\mathbf{\hat{n}}(\mathbf{q})$ are similarly defined by level sets of the probability distribution to be sampled.

Since $\tau$ only appears in the combination $\frac{\tau}{m}\mathbf{p}$, a scaling of $\tau$ is equivalent to an appropriate change in $\sigma_p$ so that we set $\tau=1$ without loss of generality.
Similarly, we take $m=1$ so that momentum, velocity and position have the same units. 
The absolute scale is set by the volume under consideration.
In the literature, $\sigma_p$ is commonly referred to as the step size of the algorithm. Time $t$ is measured in units of Monte Carlo Steps (MCS). Finally, we use the notation $\mathbf{\hat{x}}$ to denote the normalisation of the vector $\mathbf{x}$ throughout.
In summary, \gls{gmc} has three parameters, $\sigma_p$, $L$ and the number of trajectories, which must be jointly tuned in practice. 

In comparison with dynamical billiards, which is the Hamiltonian flow of a particle in a potential constant inside and infinite outside the boundary, there are two fundamental differences. Firstly, \gls{gmc} dynamics introduces additional discontinuities into the billiards flow map. 
To see this, consider a particle moving close to a flat boundary, in parallel. Considering the effective momentum of a particle, $\mathbf{p}_\mathrm{eff}=\mathbf{q}(t+1)-\mathbf{q}(t)$, it is clear that $\mathbf{p}_\mathrm{eff}=\mathbf{p}(t)$.
However, if the momentum of the particle has a small angle towards the boundary such that it crosses the boundary in the next time step, we approximately have $|\mathbf{p}_\mathrm{eff}|\approx 2|\mathbf{p}(t)|$ since the particle evolves under the reflection branch and hence takes two position updates in a single step.
Generalising to an ensemble of particles co-existing in a small neighbourhood in phase space, these discontinuities split the ensemble into spatially separated subpopulations, travelling at two different effective speeds. In the sphere and cube, we show this splitting explicitly.

Secondly, by the definition of dynamical billiards, the reflection is fixed in space and neighbouring particles reflect with a time delay determined by the momentum normal to the boundary~\cite{dellago1996lyapunov}. However, in \gls{gmc} dynamics, two particles sufficiently close in phase space reflect at the same time. In the sphere and cube, this simultaneous reflection is accompanied with bunching of the particle density, i.e. the formation of a local over-density of particles. 
Specifically, in the sphere, first note that a reflection preserves the radial position of a particle, i.e. $|\mathbf{q}(t+1)|=|\mathbf{q}(t)|$. For example, Figure~\ref{fig:supersonic-figure} shows this. 
As we will show in Section~\ref{sec:sphere-lossless-two-dimensional-representation}, for a typical initialisation, most particles will reflect in the next time step, assuming intermediate $\sigma_p$. 
Now, if we have an ensemble of particles initialised at the same position, this implies that most particles will still have the same radius at the next time step.
The dimensionality of the support of the particle distribution is therefore $\dimension-1$ since all particles are squashed onto the same radius.
As we will show, a significant fraction of the particles continue to remain at the initial position in subsequent time steps.
This illustrates the previously mentioned bunching of a particle distribution.
In contrast, for the same initialisation in dynamical billiards, all particles immediately obtain different radial positions, regardless of the value of $\sigma_p$. The particle distribution will continue to gradually spread in the full dimensionality $\dimension$ of the space. 
In the cube, we will show that a dispersing wave packet will be focused by a reflection in Section~\ref{sec:cube-low-dimensional-representation} and Figure~\ref{fig:one-dimensional-particles-sketch}, which is again not the case for billiards.
The described instances of bunching manifest as oscillations in the \gls{sd} because the \gls{sd} measures the non-uniformity of the particle distribution. This will become clear in the following sections. 
We call the general phenomenon associated with bunching a resonance.

Finally, we note that there is no Hamiltonian generating a continuous-time dynamics whose restriction to discrete times is \gls{gmc} dynamics. Alternatively formulated, we cannot extend \gls{gmc} to continuous time even if we change the potential. 
Any modification on the potential must be outside the boundary since particles travel in straight lines inside.
In one dimension, for a particle in a box of length $2\ell$, $[-\ell,\ell]$, suppose that the particle is travelling in the positive $\mathbf{q}$-direction with momentum~$\mathbf{p}$. Assuming that the \gls{gmc} Markov Chain is ergodic,
the particle will be located at position $\mathbf{q}=\ell-\mathbf{p}+\varepsilon$ at some time $t$, where we choose $0<\varepsilon<\frac{\mathbf{p}}{2}$. The positivity condition on $\varepsilon$ ensures that the particle is mapped to $\mathbf{q}$ at time~$t+1$ again because it reflects.
However, a continuous dynamics will take at least a time interval of $2\frac{\mathbf{p}-\varepsilon}{\mathbf{p}}>1$ for the particle to return to position $\mathbf{q}$, so that the particle does not make it back in time to the same position~$\mathbf{q}$. While the above argument holds in one dimension, we conjecture that \gls{gmc} cannot be extended to continuous time dynamics in higher dimensions, as well. In fact, if there were such an extension, we could take the radial dynamics in the $(\dimension-1)$-sphere as a one-dimensional continuous extension of \gls{gmc}, which we have shown is not possible. The implication of the above is that our subsequent analysis must be based solely on the dynamical equations of \gls{gmc}.

\subsection{Sinkhorn divergences}\label{sec:sinkhorn-divergences}

In this section, we define the main tool to analyse the mixing of \gls{gmc}.
As a preliminary definition, we introduce the entropy-regularised optimal transport cost~\cite{cuturi2013sinkhorn, genevay2016stochastic}, which is defined as
\begin{equation}\label{eqn:entropy-regularised-optimal-transport-cost}
	\mathrm{OT}_\varepsilon(\alpha,\beta)=\min_{\pi\in \Pi(\alpha,\beta)}\int c(x,y)
	\pi(x,y)\mathrm{d}x\mathrm{d}y
	+\varepsilon\mathrm{KL}(\pi|\alpha\otimes \beta),
\end{equation}
for any $\varepsilon>0$ and probability distributions $\alpha$ and~$\beta$ on $\mathbb{R}^\dimension$,
where $c$ is a positive symmetric cost function and 
\begin{equation}
	\mathrm{KL}(\pi|\alpha\otimes \beta)=\int\log\left(\frac{\pi(x,y)}{\alpha(x)\beta(y)}\right)\pi(x,y)\mathrm{d}x\mathrm{d}y
\end{equation}
is the Kullback-Leibler divergence between $\pi$ and the product distribution $\alpha\otimes\beta$. 
The minimisation in Equation~\ref{eqn:entropy-regularised-optimal-transport-cost} is performed over all $\pi$ in the set $\Pi(\alpha,\beta)$ of joint distributions on the support of $\alpha$ and $\beta$ with $\alpha$ and $\beta$ as marginal distributions.
In the following, we always choose ${c(x,y)=\sum_{i=1}^\dimension |x_i-y_i|}$. For $\varepsilon\rightarrow 0$, we recover the optimal transport cost between the distributions $\alpha$ and $\beta$. For $\varepsilon>0$, the entropic regularisation through the  Kullback-Leibler divergence renders the minimisation problem convex 
and it can be solved efficiently on a GPU for empirical distributions defined through samples. 

The problem with $\mathrm{OT}_\varepsilon$ is that $\mathrm{OT}_\varepsilon(\alpha,\alpha)\neq 0$. As a remedy, the \gls{sd} was introduced~\cite{genevay2018learning},
\begin{equation}\label{eqn:sinkhorn-divergence}
	\mathrm{SD}(\alpha,\beta)=\mathrm{OT}_\varepsilon(\alpha,\beta)-\frac12\mathrm{OT}_\varepsilon(\alpha,\alpha)-\frac12\mathrm{OT}_\varepsilon(\beta,\beta),
\end{equation}
which does not have this bias.

Importantly, we have $\mathrm{SD}(\alpha,\beta)\ge0$, $\mathrm{SD}(\alpha, \beta)=0$ if and only if $\alpha=\beta$ and $\mathrm{SD}(\alpha,\beta)\rightarrow 0$ if $\alpha$ converges to $\beta$ in law~\cite{feydy2019interpolating}. If we have access to $N$ samples from each of the two distributions, we can measure the divergence between the empirical distributions defined by the samples.
The accuracy of the divergence between the empirical distributions when compared to the divergence between the underlying distributions is termed the sample efficiency, which scales as $N^{-1/2}$ with a constant dependent on $\varepsilon$ and $\dimension$~\cite{genevay2019sample}.

In the following, we always choose~$\beta$ to be the empirical distribution of $N$ independent and identically distributed samples drawn from the uniform distribution on the particular volume under consideration. While it is possible to sample exactly from the sphere and cube, other volumes can generally be sampled from by rejection sampling. The samples defining $\alpha$ will evolve in time under the \gls{gmc} Markov Chain, initialised from the same position but with momenta drawn independently from the Gaussian $\mathcal{N}(\mathbf{0},\sigma_p^2\mathbf{1})$. The initialisation in configuration space is therefore a Dirac delta distribution and the decrease in \gls{sd} measures the mixing of the Markov Chain in time, while an increase indicates that $\alpha$ is becoming less uniform, or equivalently that the underlying particles exhibit bunching. For $t\rightarrow\infty$, we expect that $\mathrm{SD}(\alpha,\beta)\rightarrow 0$ due to the convergence of the Markov Chain.

In the following, we use the implementation of Equation~\ref{eqn:sinkhorn-divergence} in the \textsc{OTT} library~\cite{cuturi2022optimal}. We choose $\varepsilon=4$, such that $\mathrm{SD}\approx 0$ for two sets of uniformly distributed samples in the sphere and cube for $2\times 10^3$ samples in $\dimension=100$ dimensions. Increasing $\varepsilon$ smooths variations in the \gls{sd} when measured as a function of time, but does not change the timescales of the variations and the frequency content, which we mainly use in the following.

\section{Dynamics in a unit sphere}\label{sec:dynamics-in-unit-sphere}

\begin{figure*}
	\centering
	\includegraphics{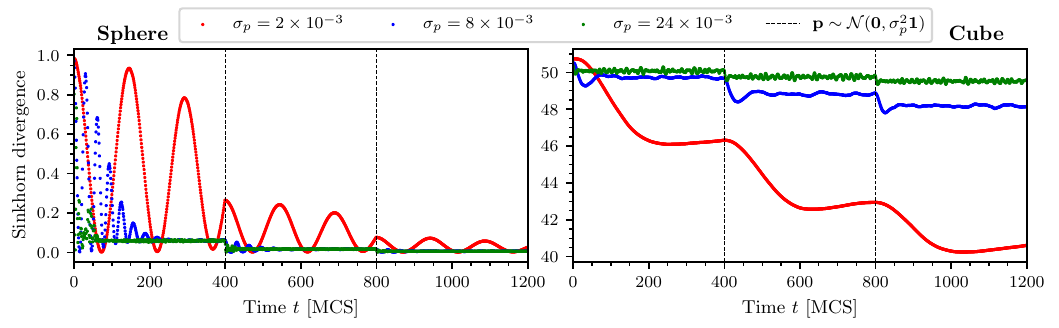}
	\caption{
	Sinkhorn divergence (SD) of an ensemble of \gls{gmc} Markov Chains in the sphere (left) and the cube (right) in $\dimension=100$ dimensions, measured with respect to the uniform distribution. Time is measured in Monte Carlo steps (MCS).  Initialised from a randomly chosen point, $10^3$ particles are evolved under \gls{gmc} dynamics with momenta drawn independently from a multivariate Gaussian $\mathcal{N}$ with standard deviation $\sigma_p$, commonly known as the step size of the algorithm. Every $L=400\,\mathrm{MCS}$, the momenta are re-randomised. 
	Within a single trajectory, i.e. up to the momentum re-randomisation, the particle ensemble converges to a subspace of the full volume so that the \gls{sd} stagnates on a timescale given by the dispersion the particle distribution. The observed oscillations are caused by temporary bunching (resonances) of the particles, both due to the inexact reflections present in the \gls{gmc} dynamics and concentration phenomena in high dimensions.
	Momentum re-randomisation effectively restarts the algorithm, leading to an approximately repeating pattern in the \gls{sd}, which  converges to zero on longer timescales since the stationary distribution of the Markov Chain is uniform in the volume.
	}
	\label{fig:long-trajectory-ball}
\end{figure*}

In this section, we consider the volume bounded by the unit sphere, $S^{\dimension-1}=\{\mathbf{q}\in\mathbb{R}^\dimension\mid \sum_{i}q_i^2=1\}$.
Since~\gls{gmc} requires gradients outside the sphere, we define a unit vector field pointing towards the origin, $\mathbf{\hat{n}}(\mathbf{q})=-\mathbf{\hat{q}}$. In nested sampling, this corresponds to the case of an isotropic distribution such as a Gaussian with isotropic covariance.

\subsection{Empirical mixing results}\label{sec:empirical-mixing-results}

The left subfigure of Figure~\ref{fig:long-trajectory-ball} shows the \gls{sd} against time for an ensemble of chains of \gls{gmc}, initialised at the same position drawn uniformly from the unit ball with momenta sampled from $\mathcal{N}(\mathbf{0},\sigma_p^2\mathbf{1})$ in $\dimension=100$ dimensions. At time $t=0$, all particles are localised at the same position, resulting in a large \gls{sd}. In high dimensions, the radius of the initial position is almost at the boundary. As the particles spread out, the \gls{sd} decreases rapidly. 
In high dimensions, the radial distribution of the Gaussian momentum distribution is concentrated at~$|\mathbf{p}|=\sigma_p\sqrt{\dimension}$ with standard deviation $\frac{\sigma_p}{\sqrt{2}}$~\cite{vershynin2018high}. The particle distribution therefore spreads like a narrow shock wave which reflects at the boundaries.
The momentum standard deviation $\sigma_p=2\times 10^{-3}$ is sufficiently small that the discretisation is negligible and the particle density behaves fluid-like. The \gls{sd} reaches a minimum before the particles reassemble into a Delta peak at approximately $-\mathbf{q}(0)$. Since the radial distribution has a finite standard deviation, this Delta peak is smeared out so that the \gls{sd} does not rise to the value at $t=0$. This process continues, resembling a fluid which oscillates back and forth between $\mathbf{q}(0)$ and $-\mathbf{q}(0)$. Re-randomising the momenta of all particles every $L=400\,\unittime$ introduces a kink in the evolution of the \gls{sd} against time. As expected, the \gls{sd} decreases to lower values at every re-randomisation since the algorithm is only ergodic due to the momentum re-randomisations. Moreover, we observe roughly the same pattern in the \gls{sd} against time at every re-randomisation, in accordance with the notion that the algorithm restarts for each particle.

When the momentum standard deviation is increased to $\sigma_p=8\times 10^{-3}$, the frequency of the oscillations increases. We observe that the \gls{sd} reaches a finite plateau, indicating that the particle distribution has converged to a subspace of the unit ball. In particular, at this point the particle distribution exhibits an under-density at the centre of the sphere, forming a hole. Meanwhile, the angular distribution is approximately uniform. The convergence speed to the subspace increases with $\sigma_p$ as this is driven by diffusive spreading of particles which is induced by the finite width of the radial momentum distribution,~$\frac{\sigma_p}{\sqrt{2}}$. Larger $\sigma_p$ therefore causes faster convergence.
Subsequent re-randomisation of the particle momenta moves each particle onto a new contour in phase space, allowing the particles to explore a larger subspace, leading to an almost discontinuous drop in the \gls{sd}. This pattern repeats until the particle distribution converges to the uniform distribution with the \gls{sd} converging to zero, as expected, since this is the stationary distribution of the Markov Chain.
Increasing $\sigma_p$ further to $24\times 10^{-3}$ agrees with the picture developed above.

Note that the value of the \gls{sd} being approximately zero, as in Figure~\ref{fig:long-trajectory-ball}, does not imply a uniform particle distribution as decreasing the regularisation parameter~$\varepsilon$ shifts the curves to larger absolute values. In fact, especially the radial distribution function exhibits little mixing at this point. We provide a detailed description of this in the following section.

\subsection{Lossless two-dimensional representation}\label{sec:sphere-lossless-two-dimensional-representation}

\begin{figure}[!]
	\centering
	\includegraphics[width=0.8\columnwidth]{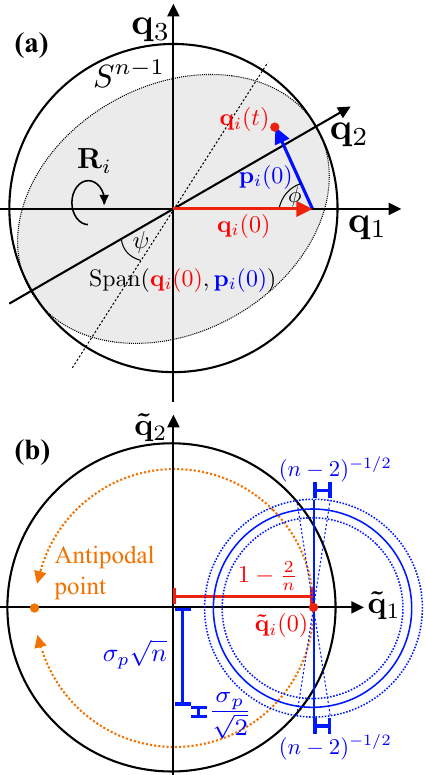}
	\caption{(a) Action of the rotation map $\mathbf{R}_i$ (Equation~\ref{eqn:disk-map}). In the sphere $S^{n-1}$, a trajectory lies in a two-dimensional disc spanned by the initial position $\mathbf{q}_i(0)$ and momentum $\mathbf{p}_i(0)$. All particles have different momenta but are initialised from the same point and hence share the axis $\mathbf{q}_i(0)$. Their individual discs can therefore be rotated onto the $\mathbf{q}_1$-$\mathbf{q}_2$-plane. This maps the angle $\psi$ to zero but preserves $\theta$. (b) Image of the rotation map. The radial and angular distributions of the initial momentum concentrate around $\sigma_p\sqrt{\dimension}\pm\frac{\sigma_p}{\sqrt{2}}$ and $\pm\frac\pi2\pm(\dimension-2)^{-1/2}$, respectively. To first order in $\frac{1}{\dimension}$, the distance of the initial position is $1-\frac2\dimension$ from the origin. The dominant motion in high dimensions therefore approximately follows a path close to the boundary, converging at an antipodal point.}
	\label{fig:action-rotation-map}
\end{figure}

We now focus on an ensemble of deterministic trajectories of \gls{gmc}, i.e. an ensemble of particles initialised at the same position with momenta drawn from $\mathcal{N}(\mathbf{0},\sigma_p^2\mathbf{1})$ but without subsequent momentum re-randomisations. This corresponds to the behaviour of the \gls{sd} in Figure~\ref{fig:long-trajectory-ball} up to the first re-randomisation time.

By considering the geometry of a reflection, it can be seen that a single particle is reflected onto a position of the same radius. In particular, this implies that the Markov Chain is rejection free, i.e. the rejection branch is never invoked.
Moreover, the motion is confined to a two-dimensional disc. This is seen by noting that the initial position vector, measured from the centre of the sphere, the momentum vector and the normal vector lie in the same plane. Hence, any subsequent positions and momenta defined by the \gls{gmc} dynamics
can be expressed as linear combinations of the initial conditions so that they remain confined in this plane.

Consequently, we can map all trajectories onto a single two-dimensional disc while preserving radial distribution functions as follows (Figure~\ref{fig:action-rotation-map}). First, assume that the initial position is aligned with the $\mathbf{q}_1$-axis. Otherwise we can perform a global rotation to achieve this. For each particle $i$, there is now a rotation matrix $\mathbf{R}_i$, constructed explicitly in Appendix~\ref{sec:appendix-explicit-construction}, and projection matrix~$\mathbf{P}_i=\begin{pmatrix}\mathbf{1}_{2\times 2}&\mathbf{0}_{2\times (\dimension-2)}\end{pmatrix}$, where~$\mathbf{1}_{2\times2}$ is the two-dimensional identity matrix and~$\mathbf{0}_{2\times (\dimension-2)}$ is a matrix of zeroes of size $2\times(\dimension-2)$, 
which map the $\dimension$-dimensional position vector $\mathbf{q}_i(t)$ to a two-dimensional vector,
\begin{equation}\label{eqn:disk-map}
	\mathbf{\tilde{q}}_i(t)=\mathbf{P}_i\mathbf{R}_i\mathbf{q}_i(t),
\end{equation} 
by first rotating the two-dimensional disc around the axis spanned by the sphere centre and the initial position, giving $\mathbf{R}_i\mathbf{q}_i(t)=((\tilde{q}_i)_1(t),(\tilde{q}_i)_2(t),0,\dots,0)^\top$, and subsequently discarding the dimensions zeroed by $\mathbf{R}_i$. In order to preserve the symmetry of isotropic distributions, we choose to rotate either clockwise or anti-clockwise such that half of the particles end up on the upper or lower half of the two-dimensional disk. We use the same transformation on the momenta, thus giving two-dimensional momenta $\mathbf{\tilde{p}}_i(t)=\mathbf{P}_i\mathbf{R}_i\mathbf{p}_i(t)$. The radial distribution of~$\mathbf{\tilde{p}}_i$ is identical to that of $\mathbf{p}_i$ as lengths are preserved under application of $\mathbf{P}_i\mathbf{R}_i$. 
While the distribution of $\mathbf{\hat{p}}_i=\mathbf{p}_i/|\mathbf{p}_i|$ is uniform, the distribution of $\mathbf{\hat{\tilde{p}}}_i=\mathbf{\tilde{p}}_i/|\mathbf{\tilde{p}}_i|$ is proportional to $|\sin\phi|^{\dimension-2}$ (Appendix~\ref{sec:appendix-derivation-circle-density}), where $\phi$ is the angle between $\mathbf{q}_i$ and $\mathbf{p}_i$. The effective angular distribution hence concentrates around $\phi=\pm\frac{\pi}{2}$. The standard deviation of $\phi$ decreases as 
$(n-2)^{-1/2}$
with increasing dimensionality $\dimension$, obtained by a Laplace approximation. 
Intuitively, the overlap of the isotropically distributed momentum vector with the $(\dimension-1)$-dimensional tangent space at the initial position is significantly larger than the overlap with the one-dimensional radial vector. Hence, in high dimensions, a particle is most likely to initially move orthogonally to its initial position.
We stress that the map in Equation~\ref{eqn:disk-map} is an effectively lossless compression of the high-dimensional particle positions onto two dimensions, enabled by the symmetries of the sphere.

The two-dimensional positions $\{\mathbf{\tilde{q}}_i(t)\}_{i=1}^N$ can be directly visualised. To avoid numerical instabilities, we do not apply Equation~\ref{eqn:disk-map} to high-dimensional particles. Instead, we use the analytically derived two-dimensional distributions to directly prepare the particles in two dimensions. To initialise the particle positions, we sample~$\mathbf{q}_i(0)$ from the $\dimension$-dimensional unit ball and set ${\mathbf{\tilde{q}}_i(0)=(|\mathbf{q}_i(0)|,0,\dots,0)}$. To initialise the momenta, we sample $\{\mathbf{p}_i(0)\}_{i=1}^N$ from the $\dimension$-dimensional normal distribution, giving $\mathbf{\tilde{p}}_i(0)$ in polar coordinates as ${|\mathbf{\tilde{p}}_i(0)|=|\mathbf{p}_i(0)|}$ and angle ${\theta=\sgn((\mathbf{p}_i)_2(0))\cos^{-1}(\mathbf{p}_i)_1(0)}$ (Equation~\ref{eqn:point-to-angle}).

\begin{figure}[!]
	\centering
	\includegraphics{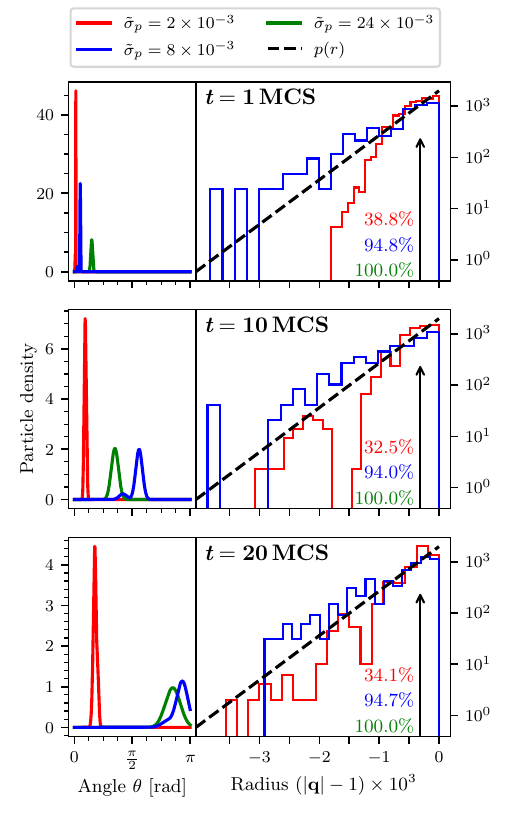}
	\caption{Angular and radial particle density for different times $t$ in the sphere for $\dimension=2\times 10^3$ and $\sigma_p=\tilde{\sigma}_p\sqrt{\frac{100}{\dimension}}$. The angle $\theta$ is defined through the rotation map (Equation~\ref{eqn:disk-map}). The dispersion and the splitting of the particle distribution is visible for $\tilde{\sigma}_p=8\times 10^{-3}$ at $t=10\,\unittime$, with the faster particles travelling at the supersonic velocity defined in Equation~\ref{eqn:supersonic-frequency} and the slower particles in a smaller wave packet behind them. At $t=20\,\unittime$, the wave packet merges at $\theta=\pi$ with the one travelling in the opposite direction. For $\tilde{\sigma}_p=24\times 10^{-3}$, the wave packet travels faster, however this motion appears slower due to aliasing. The radial distribution is effectively fixed at the initial radius, which is expected from the concentration of the angular distribution $p(\phi)\propto|\sin\phi|^{\dimension-2}$. The arrow denotes a delta function and the percentage is the fraction of particles at the delta function. For comparison, the dotted straight line is the radial distribution $p(r)=\dimension r^{\dimension-1}$ of the uniform distribution in the sphere on a log-log axis. There are no particles deeper inside the sphere than shown. This creates an under-density compared to the uniform distribution or, equivalently, an over-density near the boundary.
	}
	\label{fig:angular-and-radial-density-sphere}
\end{figure}

To show the dynamics of the particle density more clearly, we increase concentration effects by increasing the dimensionality to $\dimension=2\times 10^3$. We scale the three values of $\sigma_p$ shown in Figure~\ref{fig:long-trajectory-ball} to $\sigma_p=\tilde{\sigma}_p\sqrt{\frac{100}{\dimension}}$, where ${\tilde{\sigma}_p\in\{2\times 10^{-3},8\times 10^{-3}, 24\times 10^{-3}\}}$. This ensures that the radius of the momentum shell remains the same. 
Figure~\ref{fig:angular-and-radial-density-sphere} shows the particle density of the angle $\theta$ and radius~$|\mathbf{q}|$ as it evolves in time.
The density is a Dirac delta function at $\theta=0$ at initialisation. At $t=1\,\unittime$, the particle distributions propagate as Gaussian wave packets, as expected. The mean angle increases linearly with time, except for $\tilde{\sigma}_p=24\times 10^{-3}$. This wave packet is aliased, i.e. its effective (visible) momentum is slower than its true momentum due to the discretised motion. In addition, the angular wave packets visibly disperse and eventually spread over all angles. This diffusive component of the motion is induced by the finite radial width of the momentum distribution.

The radial distributions display a Dirac delta peak at the initial radius, as expected. The fraction of particles stuck at this radius increases as $\tilde{\sigma}_p$ is increased. For $\tilde{\sigma}_p=24\times 10^{-3}$, all particles are stuck at the same radius. Additionally, the remaining particles are confined to a narrow shell near the boundary. A comparison with the radial density, $p(r)=\dimension r^{\dimension-1}$, of the uniform distribution inside the sphere shows that there is an over-density near the boundary and a lack of particles deeper inside the sphere.

With the radius approximately fixed, all particles travel simultaneously to the antipodal point, until they meet at $-\mathbf{q}(0)$. At this point, the angle $\theta$ wraps around and we observe particles with angle larger than~$\pi$ in the range $[0,\pi]$.
The merging of the two particle distributions travelling in opposite angular directions is visible for $\tilde{\sigma}_p=8\times 10^{-3}$ at $t=20\,\unittime$.
Overall, the dynamics of the particle density supports the picture of the oscillating fluid-like motion developed in Section~\ref{sec:empirical-mixing-results}. 

In addition to the dominant mode of the angular particle density, there is a second mode of particles moving at a slower effective angular velocity. This is especially visible for $\tilde{\sigma}_p=8\times 10^{-3}$ at $t=10\,\unittime$.
We analyse the components of this motion in the following section.

\subsection{Frequency spectrum}\label{sec:dynamics-sphere-frequency-spectrum}

We model the \gls{sd} as a function of time (left subfigure of Figure~\ref{fig:long-trajectory-ball}) for a single trajectory as
\begin{equation}
	\mathrm{SD}(t)\propto \mathrm{e}^{-t/\tau_\mathrm{broad}}\sum_{i=1}^{N_\mathrm{freq}}\left[1+\cos(2\pi f_i t)\right],
\end{equation}
where $\{f_i\}_{i=1}^{N_\mathrm{freq}}$ is the frequency spectrum of density oscillations and $\tau_\mathrm{broad}$ is the timescale of decay corresponding to the decay from diffusive broadening of the wave front. The functional form of $\mathrm{e}^{-t/\tau_\mathrm{broad}}$ matches the observed exponential decay when the momentum standard deviation $\sigma_p$ is sufficiently large.
For smaller~$\sigma_p$, the decay takes a Gaussian form, $\mathrm{e}^{-t^2/2\tau^2}$. However, for the purpose of describing the dominant timescales of the dynamics, the precise functional form matters less since both functional forms lead to the same scale of broadening in frequency space, which is why we continue with the exponential decay. The power spectral density, $\mathrm{PSD}(f)\propto |\mathcal{F}[\mathrm{SD}](f)|^2$, restricted to positive frequencies is
\begin{equation}
	\mathrm{PSD}(f)\propto \left| \frac{1}{f_\mathrm{broad}^2+f^2}\star \sum_{i=1}^{N_\mathrm{freq}}[\delta(f)+\delta(f-f_i)]\right|^2,
\end{equation}
where $\mathcal{F}$ is the Fourier transform, $\star$ denotes a convolution and $f_\mathrm{broad}=\tau_\mathrm{broad}^{-1}$.

\begin{figure}
	\centering
	\includegraphics[width=\columnwidth]{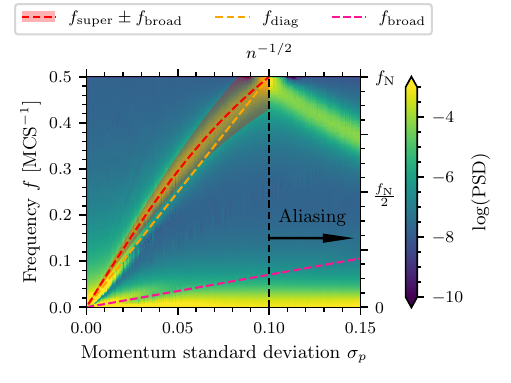}
	\caption{Power spectral density $\mathrm{PSD}(f)$ of the Sinkhorn divergence against the standard deviation $\sigma_p$ of the momentum distribution in the sphere in $\dimension=100$ dimensions.
		Increasing~$\sigma_p$ increases the average momentum magnitude of the particles, leading to higher-frequency oscillations in the particle density up to the Nyquist frequency $f_\mathrm{N}=0.5\,\mathrm{MCS}^{-1}$ when $\sigma_p\sqrt{n}\sim 1$.
		The dominant frequency $f_\mathrm{super}>f_\mathrm{diag}$ indicates that most particles travel with larger effective momentum close to the boundary. Dispersion of the particle wave front leads to broadening of the $\mathrm{PSD}$ on a scale of $f_\mathrm{broad}$.
	}
	\label{fig:unit-ball-frequency-vs-step-size}
\end{figure}

We now discuss the frequency content $\{f_i\}_{i=1}^{N_\mathrm{freq}}$. In sufficiently high dimensions, the momentum distribution is concentrated in a thin spherical shell of radius~$\sigma_p\sqrt\dimension$. We therefore expect density oscillations to originate from the propagation of the wavefront as it is reflected in the unit ball. This includes a density wave reflecting back and forth between antipodal points of the unit sphere, leading to radial oscillations of frequency 
\begin{equation}
	f_\mathrm{diag}(\sigma_p)=\frac{\sigma_p\sqrt\dimension}{2R}, 
\end{equation}
where $R=1$ is the radius of the sphere. As Figure~\ref{fig:unit-ball-frequency-vs-step-size} shows, $f_\mathrm{diag}$ provides a lower bound to the spectrum instead of constituting the dominant frequency component. In particular, the mean frequency deviates non-linearly in $\sigma_p$ from $f_\mathrm{diag}$. 
In fact, as previously shown, since the effective angular distribution 
concentrates around $\pm\frac{\pi}{2}$, we expect most particles to follow 
trajectories along the disc boundary.
Computing the frequency $f_\mathrm{super}$ along such trajectories gives approximately (Appendix~\ref{sec:appendix-derivation-of-the-supersonic-frequency}),
\begin{equation}\label{eqn:supersonic-frequency}
	f_\mathrm{super}(\sigma_p)
	=\frac{1}{\pi}
	\left(\tan^{-1}
	\frac{\sigma_p\sqrt\dimension}{R}
	+\cos^{-1}\frac{1}{\sqrt{1+
	\sigma_p^2\dimension/R^2
	}}\right),
\end{equation}
which we term the supersonic frequency since the particles move faster than particles moving in a straight line and in particular reach the antipodal point earlier despite taking a longer path along the disc boundary. Intuitively, the fact that particles have a higher effective momentum when colliding with a boundary, as explained in Section~\ref{sec:gmc}, over-compensates for the longer path.

The decay of the \gls{sd} induces a broadening of the spectrum. As the wave front has a width of approximately $\frac{\sigma_p}{\sqrt{2}}$ in momentum space, 
the timescale over which a wavefront disperses is given by $\tau_\mathrm{broad}\sim \frac{\sqrt{2}}{\sigma_p}$ so that $f_\mathrm{broad}=\tau_\mathrm{broad}^{-1}\sim\frac{\sigma_p}{\sqrt{2}}$, omitting a dimensional pre-factor of unity.

We compute the power spectral density as a function of~$\sigma_p$ in Figure~\ref{fig:unit-ball-frequency-vs-step-size}. As expected, the frequency of oscillations increases with $\sigma_p$. At $\sigma_p\sim\dimension^{-1/2}$, the Nyquist frequency $f_\mathrm{N}=\frac{1}{2}\,\unittime^{-1}$ is reached and we observe an aliased frequency spectrum for $f>f_\mathrm{N}$. This is approximately the value of $\sigma_p$ at which the particles traverse the ball in approximately a single step, $\sigma_p\sqrt\dimension\sim 1$.
For~$\sigma_p\lesssim 0.25$, we see that the dominant frequency is less than $f_\mathrm{diag}$. These are particles travelling at the ordinary non-supersonic speed, which reach the antipodal point at a later time than particles travelling diagonally through the sphere due to the longer path length. For~$\sigma_p\gtrsim 0.25$, this mode disappears and the dominant frequency becomes~$f_\mathrm{super}$, in agreement with the previously described two-dimensional representation. Additionally, see that the dominant frequencies are broadened. In particular, the zero-frequency component and $f_\mathrm{super}$ exhibit broadening on roughly the expected scale of $f_\mathrm{broad}$. A larger range of $\sigma_p$ values is shown in Appendix~\ref{sec:appendix-power-spectral-density-sphere-larger-range}.

\section{Dynamics in a cube}\label{sec:dynamics-in-cube}

In this section, we consider the $\dimension$-dimensional cube, $\mathcal{U}=[-1,1]^\dimension$, which we extend by a unit vector field ${\mathbf{\hat{n}}_i(\mathbf{q})=-\delta_{ij}}$ where $j=\arg\max_i|\mathbf{q}_i|$. In nested sampling, this setup arises as the normalised gradient field of a pyramid.

\begin{figure}
	\centering
	\includegraphics[width=\columnwidth]{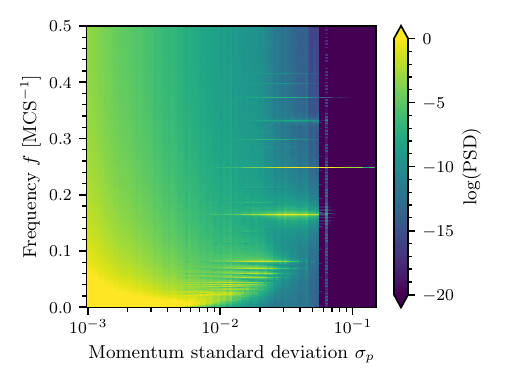}
	\caption{Transition between fluid-like ($\sigma_p\lesssim 4\times 10^{-3}$) and quasi-periodic ($\sigma_p\gtrsim 4\times 10^{-3}$) dynamics. The plot shows the power spectral density $\mathrm{PSD}(f)$ of the Sinkhorn divergence against the standard deviation $\sigma_p$ of the momentum distribution in the cube in $\dimension=100$ dimensions. At low $\sigma_p$, the particle distribution diffuses throughout the cube. Increasing $\sigma_p$ increases the rejection probability, leading to particles trapped along effectively one-dimensional trajectories, causing a discrete spectrum of resonant frequencies. At even larger $\sigma_p$, the spectral line at $\sigma_p=\frac14\,\mathrm{MCS}^{-1}$ dominates and other frequencies vanish.}
	\label{fig:power-spectral-density-cube}
\end{figure}

\subsection{Empirical mixing results}

The right subfigure of Figure~\ref{fig:long-trajectory-ball} shows the \gls{sd} as a function of time. For all $\sigma_p$, there is initially a monotonic decay of the \gls{sd}. This corresponds to the particle distribution spreading out and mixing. The \gls{sd} subsequently reaches a minimum and increases in time, corresponding to the particles unmixing. The time at which the minimum is reached corresponds to the particles hitting the boundary of the cube along their trajectory and either being reflected or reversing their momentum. Naturally, a larger $\sigma_p$ decreases this time. Moreover, the fraction of particles evolving according to the rejective branch generally increases as $\sigma_p$ is increased. The reason is that a larger step size causes larger overstepping of the boundary and therefore lowers the probability of a particle reflecting back inside the boundary. Consequently, the minimum in the \gls{sd} becomes sharper as $\sigma_p$ is increased. That is, for a sufficiently large value of $\sigma_p$, almost all particles reverse their momentum upon hitting the boundary instead of reflecting and mixing further, causing the \gls{sd} to increase almost immediately.

As the particle distribution evolves further, the \gls{sd} reaches a steady-state value and fluctuations around this value are visible. The steady-state value is larger than the minimum \gls{sd} attained previously. As we will show in Section~\ref{sec:cube-low-dimensional-representation}, the particle distribution does not continue to relax to the uniform distribution after hitting a boundary. Instead, the distribution oscillates between more spread-out and bunched states. As a measure of the non-uniformity of the particle distribution, the \gls{sd} shows this oscillation around an average non-uniform state. 

Furthermore, it is visible that the steady-state value of the \gls{sd} increases as $\sigma_p$ is increased. Two factors influence this value. Firstly, more particles are stuck at their initial position due to the increase in momentum reversals. This causes the steady-state value to increase. Additionally, more particles are trapped along one-dimensional subspaces due to momentum reversal at both ends of the subspace. We test this in Section~\ref{sec:cube-frequency-spectrum}. 
For $\sigma_p=24\times 10^{-3}$, the initial minimum is of the same magnitude as the fluctuations, showing that the particles effectively do not explore the cube beyond one-dimensional subspaces.

Overall, the support of the particle distribution therefore shrinks as $\sigma_p$ is increased, leading to a larger value of the steady-state \gls{sd}. As $\sigma_p$ is increased significantly (not shown), all particles are trapped at their initial position and the \gls{sd} remains constant in time.

Subsequent re-randomisations of the particle momenta decrease the \gls{sd} further, as expected. It is visible that the effect of a re-randomisation decreases as $\sigma_p$ is increased. Specifically, the steady-state \gls{sd} decreases more strongly for $\sigma_p=2\times 10^{-3}$ than for the other values of $\sigma_p$. This occurs because the subspace of the cube explored by a particles decreases with $\sigma_p$, as discussed above, so that the gain in exploration from a randomisation of the momentum becomes smaller for larger $\sigma_p$.

\subsection{Frequency spectrum}\label{sec:cube-frequency-spectrum}

The dynamics of a single particle in an $\dimension$-dimensional cube, $[-1,1]^\dimension$, is either quasiperiodic or periodic, depending on the relative magnitude of the components of its momentum along the coordinate axes. Moreover, as the \gls{gmc} dynamics simply reverses the momentum component along a coordinate on a reflection, the particle will retrace its positions along that coordinate. Hence, in the case of quasiperiodic motion, the dynamics is ergodic on a discrete lattice with spacings given by the momentum vector components. Similarly to the dynamics of billiards, it does not mix as the dynamics is integrable. 

Figure~\ref{fig:power-spectral-density-cube} shows the frequency spectrum of the \gls{sd} in $\dimension=100$ dimensions. The power spectral density is averaged over multiple random initialisations to ensure independence from the trajectory initial conditions. For $\sigma_p\lesssim 4\times 10^{-3}$, we observe almost continuous mixing of the particle distribution, leading to a dominant zero-frequency component broadened by the decay of the~\gls{sd}. For $\sigma_p\gtrsim 4\times 10^{-3}$, discrete frequency peaks appear. As expected from quasiperiodic motion, the spectrum of an observable contains frequencies at all integer linear combinations of the fundamental frequencies~\cite{ott2002chaos}. At ${\sigma_p\gtrsim 10^{-2}}$, the dominant frequency $\sigma_p=\frac{1}{4}\,\unittime^{-1}$ emerges and, for $\sigma_p\gtrsim 8\times 10^{-2}$, all frequencies vanish except for this frequency. For even larger $\sigma_p$ (${\sigma_p\gtrsim 0.3}$), all particles are stuck at their initial position and ${f=0\,\unittime^{-1}}$ is the only frequency present, which corresponds to a Sinkhorn divergence constant in time.

In contrast to the dynamics in the sphere, the rejection branch of \gls{gmc} is invoked. This happens particularly around the corners of the cube where a reflection may not suffice to bring the particle back inside. 
Consequently, in high dimensions, in which most of the volume of a cube is in its corners~\cite{ball1997elementary},
the fraction of rejected particles increases.

We measure the critical momentum standard deviation $(\sigma_p)_\mathrm{crit}$ above which all particles evolve under the rejection branch, and are hence fixed at the initial position for all times, by calculating the entropy of the Fourier transform of the \gls{sd}, $H=-\sum_k\hat{I}_k\log \hat{I}_k$, where $\hat{I}_k$ is obtained by normalising the power spectral density of the \gls{sd} as a function of time and the sum runs over over the frequency spectrum obtained from the discrete Fourier transform. Since $H$ measures the effective support size of a probability distribution~\cite{cover1999elements}, a large value of $H$ indicates the presence of many frequencies, whereas $H=0$ indicates that the \gls{sd} does not vary with time and hence that all particles are stuck at their initial position. The boundary between $H=0$ and $H>0$ is the critical momentum standard deviation $(\sigma_p)_\mathrm{crit}$.

\begin{figure}
	\centering
	\includegraphics{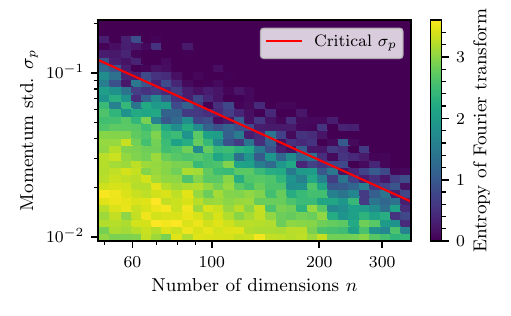}
	\caption{The transition boundary between fluid-like and quasiperiodic dynamics follows a power law in the dimension~$\dimension$, ${(\sigma_p)_\mathrm{crit}\propto n^{-0.986(2)}}$. The plot shows the entropy of the Fourier-transformed Sinkhorn divergence against the standard deviation $\sigma_p$ of the momentum distribution and the dimensionality $\dimension$. A larger value of the entropy indicates the occurrence of multiple resonances. A value of zero corresponds to the presence of a single frequency, in which case all particles are stuck due to rejections. The line indicates the critical $\sigma_p$ above which particles are rejected, where the power-law exponent is determined by an estimate of the mean chord length $\langle\ell\rangle$.}
	\label{fig:entropy-vs-ndims-step-size}
\end{figure}

Figure~\ref{fig:entropy-vs-ndims-step-size} shows the entropy $H$ as a function of~$\sigma_p$ and~$\dimension$. 
For any given value of $\dimension$, starting from $\sigma_p\approx 0$, increasing~$\sigma_p$ increases $H$ until $H$ reaches a maximum value. This is expected, since we transition from the regime dominated by the zero frequency component to the regime in which multiple discrete frequencies start to appear. Increasing $\sigma_p$ further then decreases $H$ as the diversity of the frequency spectrum decreases and only the frequency at $\sigma_p=\frac{1}{4}\,\unittime^{-1}$ remains.

Additionally, it can be seen that $(\sigma_p)_\mathrm{crit}$ decreases monotonically with~$\dimension$. 
We estimate scaling with dimension by considering the average chord length $\langle\ell\rangle$, which we define as the length of the intersection of a line with direction sampled uniformly from the sphere starting from at a point sampled uniformly from the volume, which is the cube here. 
From a Monte Carlo estimate (Appendix~\ref{sec:appendix-scaling-of-mean-chord-length}), we obtain a power-law scaling, ${\langle\ell\rangle\propto n^{-0.486\pm 0.002}}$. Since the momentum of each particle is approximately $\sigma_p\sqrt\dimension$, the critical momentum standard deviation scales roughly as ${(\sigma_p)_\mathrm{crit}\propto \langle\ell\rangle/\sqrt\dimension\propto n^{-0.986\pm 0.002}}$, which is shown in Figure~\ref{fig:entropy-vs-ndims-step-size} and aligns with the observed boundary.

\subsection{Low-dimensional representation}\label{sec:cube-low-dimensional-representation}

Since the momentum distribution is concentrated, immediately after initialisation the particle density is concentrated in a thin spherical shell which expands isotropically, akin to the propagation of a spherical shock wave. For sufficiently small $\sigma_p$, the wave front is reflected at the boundaries and simultaneously disperses, leading to a decay of the \gls{sd} with time as the particle distribution spreads out uniformly. As $\sigma_p$ is increased, an increasingly larger fraction of the particle distribution evolves according to the rejection branch. Therefore, a fraction of particles remain stuck at the initial position. The particle density is therefore a superposition of a Dirac delta distribution at the initial position and a thin non-spherical wave front. As $\sigma_p$ increases further, an increasing fraction of the freely propagating particles are rejected when they reach the boundary and are therefore confined to move back and forth along a line. 
As the radial momentum of the wave front is approximately Gaussian with mean $\sigma_p\sqrt\dimension$ and standard deviation $\frac{\sigma_p}{\sqrt{2}}$, the coarse-grained picture of the dynamics is that of a Gaussian wave packet moving along such a line.

\begin{figure*}
	\centering
	\includegraphics{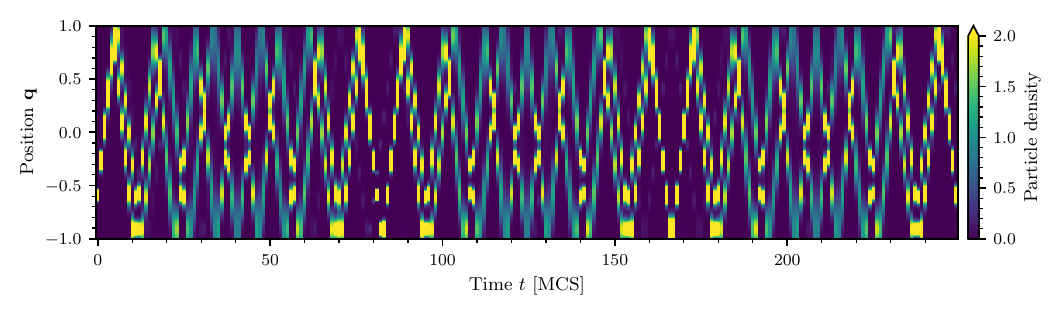}
	\caption{Particle density of a wave packet reflecting in a one-dimensional cube, $[-1, 1]$, under \gls{gmc} dynamics. The density shows periodic unmixing of the particle density caused by the inexact reflections. Additionally, the wave packet splits into two distinct wave packets visible at the same time. The particles constituting the wave packet are initialised at position $\mathbf{x}(0)=-0.9$ and have their momentum distribution set to the radial distribution of a $100$-dimensional Gaussian with $\sigma_p=3.2\times 10^{-2}$, emulating the propagation of a wave front confined to a one-dimensional line. The density is computed with a kernel density estimate from the particle distribution and clipped to $2$ (indicated by the triangle on the colour bar) to discern fluctuations in the lower density regions. The time axis takes discrete values ${t=0,1,\dots}$ but the particle density is plotted in the full intervals $[0,1],[1,2],\dots$.}
	\label{fig:one-dimensional-wave-packet-cube}
\end{figure*}

Figure~\ref{fig:one-dimensional-wave-packet-cube} shows the dynamics of a wave packet with a momentum distribution set to the radial distribution of a Gaussian in $\dimension=100$ dimensions. As the dynamics is scale-invariant in one dimension, i.e. scaling all positions $\mathbf{q}\rightarrow\lambda\mathbf{q}$ and momenta $\mathbf{p}\rightarrow\lambda\mathbf{p}$ by the same factor $\lambda$, the spacings of the lattice traced out by each particle also scale by $\lambda$, the observed density can only depend on the ratio $\frac{\mathbf{p}}{\mathbf{q}}$, where $\mathbf{q}$ and $\mathbf{p}$ are scalars here. 
We choose a one-dimensional model system on the one-dimensional cube, $[-1,1]$, having mean chord length $\langle\ell\rangle_\mathrm{1d}=2$.
Since we observe a cross-over to the fully rejective regime at $(\sigma_p)_\mathrm{crit}\approx 8\times 10^{-2}$ and the mean chord length in the $100$-dimensional cube is $\langle\ell\rangle\approx0.5$, the appropriately scaled momentum distribution takes standard deviation $(\sigma_p)_\mathrm{1d}=\langle\ell\rangle_\mathrm{1d}\sigma_p/\langle\ell\rangle=3.2\times 10^{-1}$. As we aim to observe resonances, which occur at smaller~$(\sigma_p)_\mathrm{1d}$, we decrease this value by one order of magnitude to $(\sigma_p)_\mathrm{1d}=3.2\times 10^{-2}$. The evolving particle density is plotted in Figure~\ref{fig:one-dimensional-wave-packet-cube}, which shows periodically recurring resonances in the particle density, i.e. the particle density increases locally at certain times instead of the particles spreading out uniformly over time.

\begin{figure}
	\centering
	\includegraphics{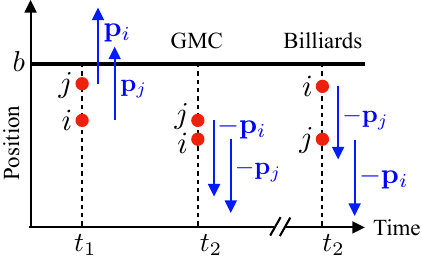}
	\caption{Inexact reflections are a cause of resonances. In high dimensions, particles move in wave packets and hence are sorted in momentum when approaching the boundary at position $b$. Particles $i$ and $j$ overstep the boundary at time $t_1$. \gls{gmc} dynamics preserves the order of particles and flips the momenta. Subsequently, the particles will converge if they were initially diverging. In dynamical billiards, the order reverses and the particles continue to diverge.}
	\label{fig:one-dimensional-particles-sketch}
\end{figure}

The resonances are a consequence of the inexact reflections (Figure~\ref{fig:one-dimensional-particles-sketch}). Tracking a pair of particles $i$ and $j$ impinging normally onto a flat boundary, the \gls{gmc} dynamics preserves the order of the particles, i.e. $\mathbf{q}_i(t_1)<\mathbf{q}_j(t_1)<b$ implies $\mathbf{q}_i(t_2)<\mathbf{q}_j(t_2)<b$ if a reflection occurs at time $t_1$ at a boundary located at $b$. This is not the case in dynamical billiards, in which the order is reversed after a reflection. As the momenta of the particles are reversed after a reflection, the particles will converge for times $t>t_2$ if they were diverging at $t<t_1$. Applying this picture to a dispersing Gaussian wave packet, in which particles are sorted by momentum, $\mathbf{p}_i(t)<\mathbf{p}_j(t)$ if $\mathbf{q}_i(t)<\mathbf{q}_j(t)$ for all particle pairs $i$ and $j$ at all times $t$ before a reflection, the wave packet becomes anti-dispersing after a reflection and is focussed to a point. This causes the observed resonances in the particle density so that the Markov Chain defined by \gls{gmc} temporarily unmixes. This does not contradict the convergence of the Markov Chain to the uniform distribution as the Markov Chain is only ergodic under repeated re-randomizations of the momentum, whereas we are discussing only a single deterministic trajectory.

The dominant frequency $f=\frac{1}{4}\,\unittime^{-1}$ appears here by considering a wave packet with a sufficiently large average momentum such that it just reaches the other end of the boundary, reflects, returns to its initial position and reflects at the other end of the boundary.
Hence, after $4$ steps, the wave packet returns to its initial position in phase space, i.e. initial position in configuration space with all particles having their initial momenta. This amounts to a period of $4\,\unittime$ and the observed dominant frequency.

Finally, we note that the wave packet is cut into multiple wave packets since each reflection induces a time delay by $1\,\mathrm{MCS}$ in which a particle remains at its current position. This explains why two wave packets are simultaneously present after a single reflection, as seen in Figure~\ref{fig:one-dimensional-wave-packet-cube}. 
Both wave packets are anti-dispersing and hence cause resonances, albeit delayed in time.
In general, the number of wave packets appearing after a reflection depends on the width of the wave packet immediately before the reflection and the momentum distribution of the wave packet so that a reflection may produce multiple wave packets.

\section{Damping resonances}

\begin{figure*}
	\centering
	\includegraphics{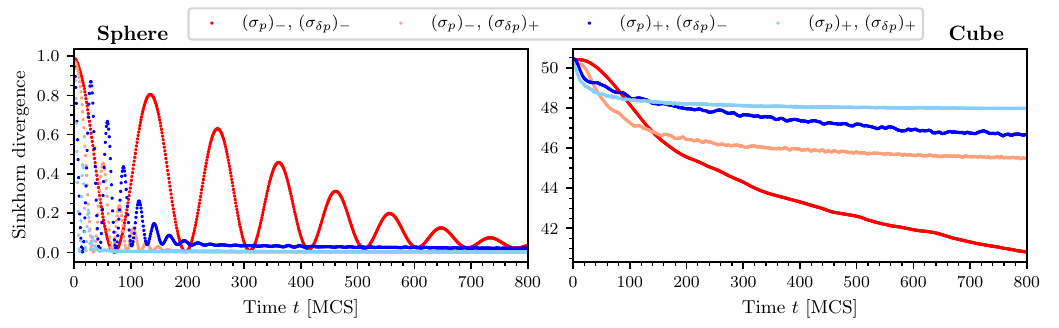}
	\caption{Adding noise to the particle momentum at every step causes resonances to decay with time. Here, the Markov Chain is initialised with a momentum $\mathbf{p}\sim \mathcal{N}(\mathbf{0},\sigma_{p}^2\mathbf{1})$ and subsequently alternates between a \gls{gmc} step and the addition of $\delta \mathbf{p}\sim \mathcal{N}(\mathbf{0},\sigma_{\delta p}^2\mathbf{1})$ to the momentum vector. We show all combinations of the values $(\sigma_p)_-=2\times 10^{-3}$, $(\sigma_p)_+=8\times 10^{-3}$, $(\sigma_{\delta p})_-=\frac{\sigma_p}{\sqrt{L}}$ and $(\sigma_{\delta p})_+=10(\sigma_{\delta p})_-$, where $L=400$. The \gls{sd} is calculated for $10^3$ independent and identically distributed Markov Chains initialised at the same position. In the sphere, the resonances (Figure~\ref{fig:long-trajectory-ball}) are still present, although increasingly damped over time. In the cube, the \gls{sd} decreases approximately linearly over longer time scales instead of plateauing. The continuous decrease stands in contrast to the discontinuities induced by full momentum re-randomisations in Figure~\ref{fig:long-trajectory-ball}.}
	\label{fig:mixing-gmc-with-noise}
\end{figure*}

Resonances can be damped by adding noise to the momentum vector at every step, see for example Algorithm~5 in~\cite{lemos2023improving}. Here, we implement the following Markov Chain. It is initialised by drawing ${\mathbf{p}\sim \mathcal{N}(\mathbf{0},\sigma_p^2\mathbf{1})}$, as before. It then proceeds by repeatedly alternating between a \gls{gmc} step and updating the momentum to $\mathbf{p}+\delta \mathbf{p}$, where $\delta \mathbf{p}\sim\mathcal{N}(\mathbf{0},\sigma_{\delta p}^2 \mathbf{1})$. Therefore, the tuning parameter $L$ is replaced by the noise strength~$\sigma_{\delta p}$. 

We note that, while this algorithm is reminiscent of underdamped Langevin dynamics, it is missing a friction term in the momentum update. As we are increasing the kinetic energy of the particle ensemble at every step, the momentum distribution does not converge to a stationary distribution. For a system of free particles, the standard deviation of the instantaneous momentum distribution diverges with time since the particles perform a random walk in momentum space. However, as we are investigating the particle dynamics of existing implementations~\cite{lemos2023improving}, we choose to focus on this algorithm.

Now, if $\sigma_{\delta p}$ is too small, the effect of the noise is negligible and the Markov Chain reduces to a single long trajectory of \gls{gmc}. On the other hand, if $\sigma_{\delta p}$ is comparable to $\sigma_p$ or too large, the Markov Chain effectively reduces to \gls{gmc} with $L=1$ and exhibits diffusive behaviour.
To provide a comparison with Figure~\ref{fig:long-trajectory-ball}, we equate the cumulative effect of noising the momentum over $L$ steps with the full re-randomisation. In particular, the average change in the momentum magnitude over $L$ steps is approximately $\sqrt{L}\sigma_{\delta p}$. This is only strictly true if we did not have reflections, which introduce correlations between particle momenta. However, we aim for only roughly similar behaviour so that we set $\sigma_{\delta p}=\frac{\sigma_p}{\sqrt{L}}$.

Figure~\ref{fig:mixing-gmc-with-noise} shows the \gls{sd} against time for this Markov Chain for $\sigma_p\in\{2\times 10^{-3},8\times 10^{-3}\}$ and the noise strengths $\sigma_{\delta p}\in\{\frac{\sigma_{p}}{\sqrt{L}}, 10\frac{\sigma_{p}}{\sqrt{L}}\}$, where $L=400$ is taken from Figure~\ref{fig:long-trajectory-ball}. 
We show the \gls{sd} up to $800\,\unittime$ as the trend in the variation of the \gls{sd} continues thereafter.

While the total decrease in \gls{sd} over $1200\,\unittime$ is slightly larger than for \gls{gmc}, we caution from concluding that the noisy Markov Chain mixes faster since we do not compute error bars on the \gls{sd} to assess statistical significance and we may further optimise the \gls{gmc} parameter $L$ in Figure~\ref{fig:long-trajectory-ball}.
Therefore, we conclude at this point that the overall decrease in the \gls{sd} over $1200\,\unittime$ for $\sigma_{\delta p}=\frac{\sigma_p}{\sqrt{L}}$ is similar to that in Figure~\ref{fig:long-trajectory-ball}, as expected from the choice of $\sigma_{\delta p}$.

In the sphere, Figure~\ref{fig:mixing-gmc-with-noise} shows that the oscillations in the \gls{sd} are damped over time and the \gls{sd} averaged over an oscillation continuously decreases, as expected. The average \gls{sd} initially decreases rapidly as in Figure~\ref{fig:long-trajectory-ball} and shows an approximately linear drift at later times. Increasing $\sigma_{\delta p}$ increases the decay rate. Moreover, larger $\sigma_{\delta p}$ increases the frequency of oscillations. This can be understood by noting that adding $\delta \mathbf{p}$ broadens the wave front of the particle ensemble so that the particle ensemble reaches the antipodal point of the sphere at an earlier time.

In the cube, the initial rapid decay of the \gls{sd} is similar to Figure~\ref{fig:long-trajectory-ball} as well. Instead of unmixing and plateauing, the \gls{sd} also exhibits an approximately linear decrease on longer time scales. That is, the resonances are dominated by the effect of the noise, as expected, since they are subdominant compared to the plateauing to begin with and less pronounced than in the sphere. 

In contrast to the sphere, increasing $\sigma_{\delta p}$ causes the Markov Chain to mix more slowly. This may be understood by noting that the mixing of \gls{gmc} is bottlenecked by high rejection rates and the entailed trapping in one-dimensional subspaces.
The comparison above with Langevin dynamics suggests that the average particle momentum increases,
which therefore increases the average speed of the particles and hence the rejection rate. 
This increases the probability of trapping in such a subspace, leading to a slower decay rate. The dynamics in the sphere, in contrast, is rejection free and hence any increase in momentum increases the decay rate, consistent with the discussion in Section~\ref{sec:empirical-mixing-results}. Overall, this suggests that the optimal value of $\sigma_{\delta p}$ depends on the dynamics in the subspace trapping the particles.

Indeed, the slower mixing with increasing $\sigma_{\delta p}$ appears contradictory with Figure~\ref{fig:long-trajectory-ball}. Specifically, suppose that we decrease $L$ to $L=50\,\unittime$ for ${\sigma_p=8\times 10^{-3}}$ in Figure~\ref{fig:long-trajectory-ball}. This would re-randomise the momentum almost exactly when the \gls{sd} converges to a plateau and hence decrease the \gls{sd} more rapidly. Decreasing $L$ corresponds to increasing $\sigma_{\delta p}=\frac{\sigma_p}{\sqrt{L}}$ in Figure~\ref{fig:mixing-gmc-with-noise}. However, as discussed above, this slows mixing down.
The contradiction is resolved by noting that the average momentum increases, as discussed, whereas \gls{gmc} fully re-randomises the momentum. Indeed, we observe in Figure~\ref{fig:mixing-gmc-with-noise} that the \gls{sd} decreases more rapidly for $\sigma_{\delta p}=10\frac{\sigma_p}{\sqrt{L}}$ than for $\sigma_{\delta p}=\frac{\sigma_p}{\sqrt{L}}$ at earlier times and only decreases more slowly at later times when the average momentum becomes significantly larger than $\sigma_p$. Re-introducing the parameter $L$ to re-randomise the momentum may prevent this, although increases the number of algorithm parameters to be jointly tuned.

In summary, the discontinuous rapid decrease of the \gls{sd} every $L$ steps in \gls{gmc} is replaced by a gradual decrease distributed over time with similar total decrease in \gls{sd}. The resonances persist on short time scales albeit decay on longer time scales. Figure~\ref{fig:mixing-gmc-with-noise} suggests that the overall decay profile of the \gls{sd} is determined by $\sigma_p$, as in \gls{gmc}, and the optimal tuning of $\sigma_{\delta p}$ faces similar considerations as for $L$, namely a dependency on the geometry of the subspace which a chain initially converges in.

Returning to \gls{gmc}, it may alternatively be possible to remove longer time-scale resonances by choosing a different momentum distribution, for example drawing the momentum direction isotropically and the magnitude from an independent Gaussian, thus preventing the concentration of the expected jump length of the Markov Chain. However, this does not remove the angular concentration present in the sphere, which is a consequence of any isotropic momentum distribution. This suggests that local preconditioning may be necessary, i.e. a coordinate transformation which depends on the current position of the particle.
Furthermore, in the context of nested sampling, we expect that the resonances can be further suppressed by increasing the number of live points.
In addition, it is an outstanding question if the average chord length, which is a global property of a particular volume, can be statistically inferred from an ensemble of live points to tune $\sigma_p$, especially in a setting in which the volume is non-convex.
While we expect that the necessary noise level and number of live points depends on the maximum height of the particle distribution on a trajectory, the joint tuning of these parameters is beyond the scope of this work.

\section{Discussion}\label{sec:discussion}

We have shown that, for the sphere and cube, an ensemble of particles initialised from the same position with a Gaussian momentum distribution evolving under \gls{gmc} dynamics exhibits resonance phenomena, which manifest as oscillations in the \gls{sd}. 
Additionally, we have shown that the relevant length scale of the sampling problem decreases as a power law in the dimension $\dimension$ for both the sphere and cube, namely $n^{-1/2}$ and $n^{(-9.86\pm 0.02)\times 10^{-1}}$, respectively. In the cube, the scale is set by the average chord length.
The persistence of the oscillations on long timescales is caused by the concentration of the Gaussian momentum distribution combined with the Dirac delta initialisation, which lead to an evolution of the particle density akin to a shock wave and hence reverberation of the particle density. On short time-scales, inexact reflections induce bunching, causing additional oscillations. 
As a consequence, the Markov Chain defined by \gls{gmc} dynamics does not mix monotonically in time but can temporarily unmix, especially in the many-chain and short-time regime. We stress that, even in the few-chain regime, the errors are still prevalent in nested sampling as errors accumulate over multiple nested sampling iterations.

In particular, the dynamics in the sphere shows that the mixing in the radial coordinate is significantly slower than in the angular coordinates. This behaviour is expected on general, non-spherical, boundaries in the high-dimensional limit, as the average length of the momentum vector projection onto the $(\dimension-1)$-dimensional tangent plane is significantly larger than the one-dimensional radial direction.
In the context of nested sampling, this biases individual samples to be too close to the boundary, therefore 
causing a negative systematic error in the Bayesian model evidence, or equivalently the partition function, which grows with dimension.
This is seen, for example in Figure~12 in~\cite{lemos2023improving}, wherein the volume at each iteration of nested sampling is non-spherical. This indicates indirectly that the results for the sphere transfer to the more general setting of ellipsoidal volumes.

Moreover, current state-of-the-art nested sampling codes, e.g.~\cite{handley2015polychord}, bring the volume at the current iteration in isotropic position, i.e. transform the coordinates such that the covariance of the uniform distribution is the identity, which is a form of preconditioning. This transformation is motivated by the fact that for other Markov Chains used on uniform distributions, such as hit-and-run slice sampling or Metropolis-Hastings with a Gaussian proposal, the spherical case is optimal. More precisely, the mixing time of hit-and-run slice sampling~\cite{lovasz1999hit, lovasz2004hit} depends on the square of a condition number, which is the ratio of radii of the smallest ball containing the volume and the largest ball inside the volume. This condition number becomes unity in the spherical case. 

To understand why the spherical case should also be the optimal case for \gls{rhmc}, we first note that the fraction of rejected points increases as the condition number of the ellipsoid increases while keeping the step size fixed. For illustration, see the depiction of the rejective branch in Figure~\ref{fig:gmc-figure}. This inherently slows mixing down. Consequently, we need to decrease the step size below the spherical case in practice, which causes the propagation of the particles to slow down. Complementary to this picture, we can also analyse the diffusive limit of \gls{rhmc} as follows.
In the limit of a single step per trajectory ($L=1$), \gls{rhmc} reduces to Metropolis-Hastings with a Gaussian proposal distribution. Taking additionally the zero step size limit ($\sigma_p\rightarrow 0$),
we obtain Brownian motion. 
In such a case, the evolution of the particle ensemble is described by the diffusion equation, $\partial_tp(\mathbf{x},t)=D\nabla^2p(\mathbf{x},t)$, with reflective boundary conditions, where $p$ is the particle density and $D$ is a positive scalar diffusion coefficient. For a Dirac delta initialisation, $p(\mathbf{x},0)=\delta(\mathbf{x}-\mathbf{x}_0)$, the solution is formally given by
\begin{equation}
	p(\mathbf{x},t)=\mathrm{e}^{Dt\nabla^2}\delta(\mathbf{x}-\mathbf{x}_0).
\end{equation}
Introducing the eigenfunctions $f_i$ of the Laplacian with reflective (Neumann) boundary conditions, ${\nabla^2f_i(\mathbf{x})=-\mu_if_i(\mathbf{x})}$, where $\mu_i\ge 0$ is the corresponding eigenvalue, the solution can be expanded as
\begin{equation}
	p(\mathbf{x},t)=p(\mathbf{x},\infty)+\sum_{i=2}^\infty\mathrm{e}^{-Dt\mu_i}f_i(\mathbf{x}_0)f_i(\mathbf{x}),
\end{equation}
where $p(\mathbf{x},\infty)$ is the steady-state uniform distribution. We see that the spectral gap of Brownian motion, $\mu_2$, dominates the decay of the initial condition and the largest possible value among all domains for a given volume is attained for a sphere~\cite{weinberger1956isoperimetric}. Hence, mixing in the sphere is fastest.

In practice, $\sigma_p$ is tuned adaptively based on a tuning metric. As the sole measurable quantity which is informative about the environment of the particle is whether a proposed point lies inside or outside the volume, tuning metrics for~\gls{rhmc} are based on tracking this quantity along particle trajectories and computing $\sigma_p$ from this information~\cite{partay2021nested, lemos2023improving}.
A example method used in practice to tune \gls{gmc} is the trajectory-wise acceptance rate~\cite{partay2021nested}, which is the fraction of points (including proposed points in the reflection and rejection branches) along a trajectory which lie inside the volume. 
Note that this acceptance rate is distinct from the one used in Hamiltonian Monte Carlo for step-size tuning, which is the acceptance rate of a single proposed point, which is always unity for rejection-free samplers such as~\gls{gmc} and consequently an uninformative metric.

We argue that the trajectory-wise acceptance rate is uninformative about mixing. Regardless of the value computed in a Markov Chain, it is not possible to detect if the Markov Chain has converged to a subspace of the volume and stopped mixing or to what extent resonances are present in the dynamics.
However, to establish at least a necessary condition for stationarity of the Markov Chain, we calculate the steady-state value of this metric by preparing independent samples in the volume and running an independent chain from each point. Since the uniform distribution is stationary under the Markov Chain~\cite{supp}, the overall particle distribution remains uniform for any choice of $\sigma_p$ so that the ensemble average of the trajectory-wise acceptance rate is equal to the time average under a Markov Chain sampling the stationary distribution. Taking the contrapositive of this statement, if a particular Markov Chain does not achieve this acceptance rate, it is not stationary. 

For the sphere, we find that the acceptance rate tends to $\frac12$ for large~$\sigma_p$. This is expected according to the previously developed picture because most particles move along the boundary so that proposed steps alternate between stepping outside and inside the volume. In the limit of $\sigma_p\rightarrow0$, the acceptance rate tends to unity, as it takes multiple steps to reach the boundary. As the dimensionality of the sphere is increased, the threshold speed is lowered above which the acceptance rate becomes $\frac12$.

For the cube, above the critical value of $\sigma_p$, individual particles become confined to one-dimensional paths. As~$\sigma_p$ is increased further, the number of steps taken in a path shrinks to $1$ so that all particles are trapped at their initial position with the momentum vector inverting at each step. Since the particles thus step outside at every third step, the limiting trajectory-wise acceptance rate takes a value of $\frac13$.

In general, for any boundary, the minimum possible value for this metric is $\frac13$. As $\sigma_p$ is decreased, the value is expected to increase. Hence, a value of at least $\frac13$ must be the tuning target, which stands in tension with the current tuning heuristic of $0.25-0.5$~\cite{partay2021nested}.

\section{Conclusions}\label{sec:conclusions}

We have provided a precise picture of the high-dimensional dynamics under inexact reflections and elucidated the mechanism underlying slow mixing, explaining long-standing practical problems with the algorithm, such as systematically negative errors in model evidences and partition functions.
In particular, we showed that the particle distribution exhibits a transition between fluid-like and discretisation-dominated behaviour. The boundary momentum standard deviation $\sigma_p$ between these regimes scales as a power law in the dimension, where the exponent depends on the shape of the volume under consideration. Moreover, the particle distribution exhibits spontaneous unmixing, regardless of the regime the algorithm operates in.
While the Markov Chain eventually converges to the desired uniform distribution under multiple momentum re-randomisations, current tuning metrics, such as the trajectory-wise acceptance rate, are disconnected from and therefore uninformative about the short-time dynamics of the algorithm which governs the mixing behaviour in the setting under consideration.
We therefore envision that our work motivates a revision of current tuning practices and provides a foundation for the development of informative tuning metrics and further piecewise-deterministic algorithms operating successfully in high dimensions.

\section*{Acknowledgements}

N. K. was supported by the Harding Distinguished Postgraduate Scholarship.
This work was performed using the Cambridge Service for Data Driven Discovery (CSD3), part of which is operated by the University of Cambridge Research Computing on behalf of the STFC DiRAC HPC Facility (www.dirac.ac.uk). The DiRAC component of CSD3 was funded by BEIS capital funding via STFC capital grants ST/P002307/1 and ST/R002452/1 and STFC operations grant ST/R00689X/1. DiRAC is part of the National e-Infrastructure.

\appendix

\section{Explicit construction of the rotation map}\label{sec:appendix-explicit-construction}

\begin{figure}
	\centering
	\includegraphics[width=0.7\columnwidth]{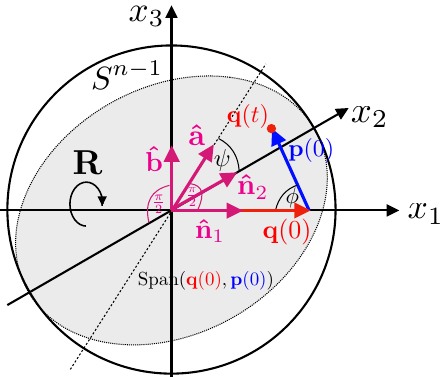}
	\caption{Vectors used in the construction of the rotation map for the choice $\mathbf{n}_2=(0,1,0,\cdots,0)^\top$.}
	\label{fig:rotation-map-construction}
\end{figure}

In this Appendix, we first construct basis vectors which then allow us to write down an explicit form of the rotation map (Figure~\ref{fig:rotation-map-construction}). 

We are given the initial position $\mathbf{q}(0)$ and $\mathbf{p}(0)$, which span the disk in which all points of a trajectory, $\{\mathbf{q}(t)\}_t$, lie.
Let $\mathbf{\hat{n}}_1$ be the unit vector parallel to $\mathbf{q}(0)$. Pick an arbitrary unit vector $\mathbf{\hat{n}}_2$ orthogonal to $\mathbf{\hat{n}}_1$. The plane spanned by $\mathbf{\hat{n}}_1$ and $\mathbf{\hat{n}}_2$ defines the plane into which the trajectory will be rotated. 
Moreover, let $\mathbf{\hat{a}}$ be the unit vector produced by Gram-Schmidt orthonormalization of $\mathbf{\hat{n}}_1$ and $\mathbf{p}(0)$, i.e. $\mathbf{\hat{a}}=\mathbf{a}/|\mathbf{a}|$ with $\mathbf{a}=(\mathbf{1}-\mathbf{\hat{n}}_1\mathbf{\hat{n}}_1^\top)\mathbf{p}(0)$. The vectors $\mathbf{\hat{n}}_1$ and $\mathbf{\hat{a}}$ form a basis for the plane of the trajectory. 
We now construct a rotation matrix $\mathbf{r}(\psi)$ which performs rotations in the $\mathbf{\hat{n}}_2$-$\mathbf{\hat{a}}$ plane with angle $\psi$. Let $\mathbf{\hat{b}}$ be the unit vector obtained by Gram-Schmidt orthonormalization of $\mathbf{\hat{n}}_2$ and $\mathbf{\hat{a}}$, i.e. $\mathbf{\hat{b}}=\mathbf{b}/|\mathbf{b}|$ with $\mathbf{b}=(\mathbf{1}-\mathbf{\hat{n}}_2\mathbf{\hat{n}}_2^\top)\mathbf{\hat{a}}$. Then, 
\begin{equation}
\mathbf{r}(\psi)=\mathbf{1}+(\mathbf{\hat{b}}\mathbf{\hat{n}}_2^\top-\mathbf{\hat{n}}_{2}\mathbf{\hat{b}}^\top)\sin\psi+(\mathbf{\hat{b}}\mathbf{\hat{b}}^\top+\mathbf{\hat{n}}_2\mathbf{\hat{n}}_2^\top)(\cos\psi-1)
\end{equation}
is a rotation matrix which reduces to the two-dimensional rotation matrix in the plane spanned by $\mathbf{\hat{n}}_2$ and $\mathbf{\hat{b}}$ and acts as the identity on any vector orthogonal to this plane. Now, let $\psi_1$ and $\psi_2$ be the angles which rotate $\mathbf{\hat{a}}$ into $\mathbf{\hat{n}}_2$ and $-\mathbf{\hat{n}}_2$, respectively. That is, $\mathbf{r}(\psi_1)\mathbf{\hat{a}}=\mathbf{\hat{n}}_2$ and $\mathbf{r}(\psi_2)\mathbf{\hat{a}}=-\mathbf{\hat{n}}_2$. We can now define the rotation matrix in Equation~\ref{eqn:disk-map} as
\begin{equation}
	\mathbf{R}=
	\begin{cases}
			\mathbf{r}(\psi_1),&\text{if $\mathbf{\hat{a}}\cdot\mathbf{\hat{n}}_2>0$,}\\
			\mathbf{r}(\psi_2),&\text{if $\mathbf{\hat{a}}\cdot\mathbf{\hat{n}}_2<0$.}
		\end{cases}
\end{equation}
Further, we can rewrite the two conditions in terms of $\mathbf{p}(0)$ as $\mathbf{\hat{a}}\cdot\mathbf{\hat{n}}_2=\mathbf{p}(0)\cdot\mathbf{\hat{n}}_2$ since $\mathbf{\hat{n}}_1\cdot\mathbf{\hat{n}}_2=0$. 

Splitting the rotation matrix $\mathbf{R}$ into two cases as above is necessary to ensure that isotropic probability distributions are mapped with equal probability into the upper and lower half-disk, respectively. Concretely, the distribution in Equation~\ref{eqn:theta-density-0} is symmetric about the $\mathbf{\hat{n}}_1$-axis.

The matrix $\mathbf{R}$ maps the $\mathbf{\hat{n}}_1$-$\mathbf{\hat{a}}$ plane onto the $\mathbf{\hat{n}}_1$-$\mathbf{\hat{n}}_2$ plane but leaves $\mathbf{\hat{n}}_1$ fixed. To wit, for any vector ${\mathbf{q}(t)=q_{1}(t)\mathbf{\hat{n}}_1+\sqrt{\mathbf{q}^2(t)-q_{1}^2(t)}\mathbf{\hat{a}}}$ in the plane of the trajectory, we have
\begin{equation}
	\mathbf{R}\mathbf{q}(t)=
	q_{1}(t)\mathbf{\hat{n}}_1+\sgn(\mathbf{p}(0)\cdot\mathbf{\hat{n}}_2)\sqrt{\mathbf{q}^2(t)-q_{1}^2(t)}\mathbf{\hat{n}}_2,
\end{equation}
as required, where $\sgn$ is the sign function. In particular, the angle $\theta$ between $\mathbf{\hat{n}}_1\propto\mathbf{q}(0)$ and $\mathbf{R}\mathbf{q}(t)$ is ${\theta=\sgn(\mathbf{p}(0)\cdot\mathbf{\hat{n}}_2)\cos^{-1}q_{1}(t)}$.

The above construction simplifies significantly for specific choices of the axes.
Without loss of generality, assume that the initial position of each particle lies on the $x_1$-axis so that $\mathbf{\hat{n}}_1\propto {\mathbf{q}(0)=(q(0),0,\dots,0)^\top}$. If this is not the case, there is a global rotation of all particle trajectories which achieves this as they are initialised at the same position. 
Now, choose $\mathbf{\hat{n}}_2=(0,1,0,\dots,0)^\top$ to be the $x_2$-axis and relabel
$\mathbf{x}=\mathbf{q}(t)=(x_1,\dots,x_\dimension)$.
Then the expression for the angle $\theta$ simplifies to
\begin{equation}\label{eqn:point-to-angle}
	\theta(\mathbf{x})=\sgn(x_2)\cos^{-1}x_1.
\end{equation}
Finally, we remark that if $\mathbf{x}$ lies on a sphere, then it is sent to the circle since $\mathbf{R}$ is a rotation.

We can also consider the action of $\mathbf{R}$ on the initial momentum $\mathbf{p}(0)$ in a coordinate system centred at $\mathbf{q}(0)$. The angle $\phi$ between $\mathbf{p}(0)$ and $\mathbf{q}(0)$ is the analogue of~$\theta$ (Figure~\ref{fig:rotation-map-construction}). In particular, $\phi$ also has a distribution ${\propto|\sin\phi|^{\dimension-2}}$ if $\mathbf{p}(0)$ is distributed isotropically (Appendix~\ref{sec:appendix-derivation-circle-density}).

\section{Derivation of the induced probability density on the circle}\label{sec:appendix-derivation-circle-density}

In the following, we show that a uniform density ${p(\mathbf{x})=1/|S^{n-1}|}$ on the unit $(n-1)$-sphere $S^{n-1}$ with surface area $|S^{n-1}|$ is mapped by Equation~\ref{eqn:disk-map} to the density
\begin{equation}\label{eqn:theta-density-0}
	p(\theta)=\frac{1}{2}\frac{\vert S^{n-2}\vert}{\vert S^{n-1}\vert}|\sin\theta|^{n-2}
\end{equation}
on the circle $S^1$ parameterized by the angle $\theta\in[-\pi,\pi]$.

For a point $\mathbf{x}$ sampled uniformly from $S^{n-1}$, the map is summarized by $\theta(\mathbf{x})=\sgn(x_2)\cos^{-1}x_1$ (Equation~\ref{eqn:point-to-angle}), where $\sgn(x_2)=x_2/|x_2|$ is the sign function and $\cos^{-1}$ is the inverse cosine with codomain $[0,\pi]$. That is, $\mathbf{x}$ is sent to the upper or lower half-circle depending on the sign of its $x_2$ component.

The induced density on the circle is therefore
\begin{equation}\label{eqn:theta-density-1}
	p(\theta)=\int\mathrm{d}S^{n-1}p(\mathbf{x})\delta(\theta-\sgn(x_2)\cos^{-1}x_1),
\end{equation}
where the integral is taken over $S^{n-1}$ and $\mathrm{d}S^{n-1}$ is the surface area element of $S^{n-1}$ such that~${|S^{n-1}|=\int\mathrm{d}S^{n-1}}$.

To proceed, we split the density $p(\mathbf{x})$ at the plane ${x_2=0}$,
\begin{equation}
	p(\mathbf{x})=\frac{1}{|S^{n-1}|}\left[\Theta(x_2)+\Theta(-x_2)\right],
\end{equation}
where $\Theta$ is the Heaviside step function. Inserting into Equation~\ref{eqn:theta-density-1} splits the integral over the sphere into two half-spheres, each with a definite value of $\sgn x_2$:
\begin{eqnarray}\label{eqn:theta-density-2}
	p(\theta)=
	&&\frac{1}{\vert S^{n-1}\vert }\left[\int\mathrm{d}S^{n-1}\Theta(x_2)\delta(\theta-\cos^{-1}x_1)\right.\\
	&&\left.+\int\mathrm{d}S^{n-1}\Theta(-x_2)\delta(\theta+\cos^{-1}x_1)
	\right]\nonumber
\end{eqnarray}

In the first integral, substituting $x_2\rightarrow -x_2$ changes the domain of integration from the half-sphere $x_2>0$ to the half-sphere $x_2<0$ but leaves the area element $\mathrm{d}S^{n-1}$ invariant, thus giving
\begin{eqnarray}
	&&\int\mathrm{d}S^{n-1}\Theta(x_2)\delta(\theta-\cos^{-1}x_1)\\
	&&=\int\mathrm{d}S^{n-1}\Theta(-x_2)\delta(\theta-\cos^{-1}x_1).\nonumber
\end{eqnarray}
Adding both sides of this Equation and using ${\Theta(-x_2)+\Theta(x_2)=1}$, we obtain
\begin{eqnarray}
	&&\int\mathrm{d}S^{n-1}\Theta(x_2)\delta(\theta-\cos^{-1}x_1)\\
	&&=\frac{1}{2}\int\mathrm{d}S^{n-1}\delta(\theta-\cos^{-1}x_1),\nonumber
\end{eqnarray}
as expected, since mapping a half-sphere onto the upper half-circle gives the same density as mapping the entire sphere onto the upper half-circle and dividing by two.

We can now evaluate this integral directly. Splitting $S^{n-1}$ along the $x_1$-axis into $(n-2)$-dimensional spheres of radius $\sqrt{1-x_1^2}$, each with arc length $\mathrm{d}x_1/\sqrt{1-x_1^2}$, we have~\cite{muller2012analysis}
\begin{equation}
	\mathrm{d}S^{n-1}=(1-x_1^2)^{(n-3)/2}\mathrm{d}x_1\mathrm{d}S^{n-2}.
\end{equation}
Moreover, using the relation 
\begin{equation}
	\delta(\theta-\cos^{-1}x_1)=\sqrt{1-x_1^2}\delta(x_1-\cos\theta) 
\end{equation}
yields
\begin{eqnarray}
	&&\int\mathrm{d}S^{n-1}\delta(\theta-\cos^{-1}x_1)\\
	&&=|S^{n-2}|\int_{-1}^1\mathrm{d}x_1(1-x_1^2)^{(n-2)/2}\delta(x_1-\cos\theta)\\
	&&=|S^{n-2}||\sin\theta|^{n-2}\Theta(\theta),
\end{eqnarray}
where the Heaviside step function was inserted to ensure that the integral is only non-zero for $\theta\in [0,\pi]$.

Similarly, the second integral in Equation~\ref{eqn:theta-density-2} evaluates to the same expression but with $\theta\rightarrow -\theta$, which holds for $\theta\in [-\pi,0]$. Substituting both integrals into Equation~\ref{eqn:theta-density-2} and using $\Theta(\theta)+\Theta(-\theta)=1$, we finally obtain Equation~\ref{eqn:theta-density-0}. 

We note that Equation~\ref{eqn:theta-density-0} is correctly normalised
since $|S^{n-2}|/|S^{n-1}|=(n-2)!!/[\kappa_n(n-3)!!]$ and $\int_0^{2\pi}\mathrm{d}\theta|\sin\theta|^{n-2}=2\kappa_n(n-3)!!/(n-2)!!$, where $\kappa_n=\pi$ if $n$ is even and $\kappa_n=2$ if $n$ is odd.

\section{Derivation of the supersonic frequency}\label{sec:appendix-derivation-of-the-supersonic-frequency}

\begin{figure}
	\centering
	\includegraphics[width=0.7\columnwidth]{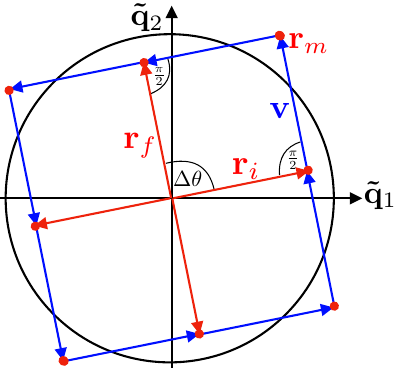}
	\caption{Sketch of the trajectory for the derivation of the supersonic frequency. Magnitude of the momentum $\mathbf{v}$ is significantly exaggerated for visualisation purposes. Here, $\mathbf{\tilde{q}}_1$ and $\mathbf{\tilde{q}}_2$ are a basis spanning the plane of the trajectory.}
	\label{fig:supersonic-figure}
\end{figure}

Assume a particle is at initial position $\mathbf{r}_i$ (Figure~\ref{fig:supersonic-figure}).
A trajectory which follows the boundary as closely as possible alternates between stepping into and out of the disk. We assume that the momentum is orthogonal to the particle position as this is the most probable configuration in high dimensions, according to Equation~\ref{eqn:theta-density-0}. After one Monte Carlo time step, the new particle position is ${\mathbf{r}_f=\mathbf{r}_i+2(\mathbf{1}-\mathbf{r}_m\mathbf{r}_m^\top)\mathbf{v}}$, where ${\mathbf{\hat{r}}_m=(\mathbf{r}_i+\mathbf{r}_f)/|\mathbf{r}_i+\mathbf{r}_f|}$. 
Since ${\mathbf{r}_i\cdot\mathbf{v}=0}$, ${\mathbf{r}_i\cdot\mathbf{\hat{r}}_m=r_0\cos(\Delta\theta/2)}$ and ${\mathbf{\hat{r}}_m\cdot\mathbf{v}=|\mathbf{v}|\sin(\Delta\theta/2)}$,
the traversed angle $\Delta\theta=\cos^{-1}(\mathbf{\hat{r}}_i\cdot \mathbf{\hat{r}}_f)$ satisfies
\begin{equation}
	\cos(\Delta\theta)=1-\frac{|\mathbf{v}|}{r_0}\sin(\Delta\theta),
\end{equation}
which yields
\begin{equation}\label{eqn:angle-traversed-supersonic-trajectory}
	\Delta\theta=\tan^{-1}
	\frac{|\mathbf{v}|}{r_0}
	+\cos^{-1}
	\frac{1}{\sqrt{1+|\mathbf{v}|^2/r_0^2}}
	,
\end{equation}
choosing the branch for which ${\Delta\theta\in[0,\pi/2]}$.

The frequency with which such a particle moves between antipodal points is approximately $\Delta\theta/\pi$, assuming that the discretisation error is negligible.
In high dimensions, the momentum distribution is concentrated in a thin shell around $|\mathbf{v}|\approx\sigma_p\sqrt\dimension$ and the initial position is close to the boundary, $r_0\approx R$. Substituting these into Equation~\ref{eqn:angle-traversed-supersonic-trajectory} gives the supersonic frequency $f_\mathrm{super}$ (Equation~\ref{eqn:supersonic-frequency}).

\section{Power spectral density for the sphere}\label{sec:appendix-power-spectral-density-sphere-larger-range}

\begin{figure}
	\centering
	\includegraphics{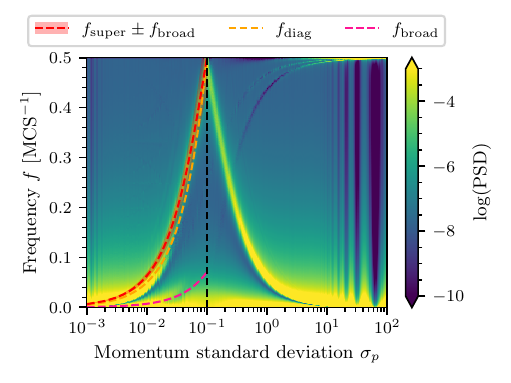}
	\caption{Same as Figure~\ref{fig:unit-ball-frequency-vs-step-size} but with a larger range of $\sigma_p$ on a log-scale.}
	\label{fig:power-spectral-density-sphere-larger-range}
\end{figure}

Figure~\ref{fig:power-spectral-density-sphere-larger-range} is identical to Figure~\ref{fig:unit-ball-frequency-vs-step-size} but includes a wider range of $\sigma_p$ values. The aliasing discussed in Section~\ref{sec:dynamics-sphere-frequency-spectrum} is more clearly visible. At $\sigma_p$ significantly larger than~$\dimension^{-1/2}$, the frequency reaches the Nyquist limit. This is a consequence of almost all particles switching between the initial and antipodal position at every time step. Therefore, the Sinkhorn divergence remains at its maximum value and no mixing is observed.

\section{Scaling of the mean chord length with dimension}\label{sec:appendix-scaling-of-mean-chord-length}

\begin{figure}
	\centering
	\includegraphics{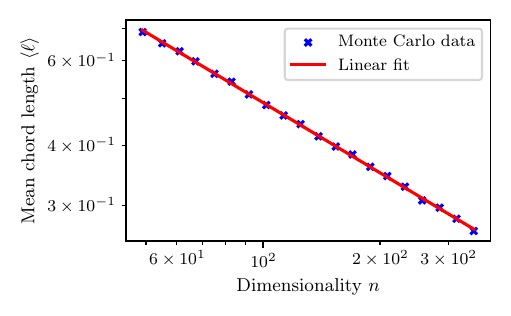}
	\caption{Monte Carlo estimate of the mean chord length in the cube for dimensionalities $\dimension\in[50, 350]$ on a log-log plot.}
	\label{fig:monte-carlo-estimate-mean-chord-length}
\end{figure}

For a given direction $\mathbf{\hat{n}}$ and point $\mathbf{x}_0$, the chord length~$\ell$ is the length of the intersection of the line ${\mathbf{x}(s)=\mathbf{x}_0+s\mathbf{\hat{n}}}$ with the cube.
The vector $\mathbf{\hat{n}}$ and $\mathbf{x}_0$ are sampled uniformly from a sphere and the cube, respectively. This gives $10^4$ samples of $\ell$, from which we compute the mean~$\langle\ell\rangle$. The dependence on dimension is shown in Figure~\ref{fig:monte-carlo-estimate-mean-chord-length} and is described by a power law with exponent $(-4.86\pm 0.02)\times 10^{-1}$.

\nocite{*}

\bibliography{gmc_problems}

\begin{thebibliography}{52}%
\makeatletter
\providecommand \@ifxundefined [1]{%
 \@ifx{#1\undefined}
}%
\providecommand \@ifnum [1]{%
 \ifnum #1\expandafter \@firstoftwo
 \else \expandafter \@secondoftwo
 \fi
}%
\providecommand \@ifx [1]{%
 \ifx #1\expandafter \@firstoftwo
 \else \expandafter \@secondoftwo
 \fi
}%
\providecommand \natexlab [1]{#1}%
\providecommand \enquote  [1]{``#1''}%
\providecommand \bibnamefont  [1]{#1}%
\providecommand \bibfnamefont [1]{#1}%
\providecommand \citenamefont [1]{#1}%
\providecommand \href@noop [0]{\@secondoftwo}%
\providecommand \href [0]{\begingroup \@sanitize@url \@href}%
\providecommand \@href[1]{\@@startlink{#1}\@@href}%
\providecommand \@@href[1]{\endgroup#1\@@endlink}%
\providecommand \@sanitize@url [0]{\catcode `\\12\catcode `\$12\catcode
  `\&12\catcode `\#12\catcode `\^12\catcode `\_12\catcode `\%12\relax}%
\providecommand \@@startlink[1]{}%
\providecommand \@@endlink[0]{}%
\providecommand \url  [0]{\begingroup\@sanitize@url \@url }%
\providecommand \@url [1]{\endgroup\@href {#1}{\urlprefix }}%
\providecommand \urlprefix  [0]{URL }%
\providecommand \Eprint [0]{\href }%
\providecommand \doibase [0]{https://doi.org/}%
\providecommand \selectlanguage [0]{\@gobble}%
\providecommand \bibinfo  [0]{\@secondoftwo}%
\providecommand \bibfield  [0]{\@secondoftwo}%
\providecommand \translation [1]{[#1]}%
\providecommand \BibitemOpen [0]{}%
\providecommand \bibitemStop [0]{}%
\providecommand \bibitemNoStop [0]{.\EOS\space}%
\providecommand \EOS [0]{\spacefactor3000\relax}%
\providecommand \BibitemShut  [1]{\csname bibitem#1\endcsname}%
\let\auto@bib@innerbib\@empty
\bibitem [{\citenamefont {Neal}(2003)}]{neal2003slice}%
  \BibitemOpen
  \bibfield  {author} {\bibinfo {author} {\bibfnamefont {R.~M.}\ \bibnamefont
  {Neal}},\ }\bibfield  {title} {\bibinfo {title} {Slice sampling},\
  }\href@noop {} {\bibfield  {journal} {\bibinfo  {journal} {The annals of
  statistics}\ }\textbf {\bibinfo {volume} {31}},\ \bibinfo {pages} {705}
  (\bibinfo {year} {2003})}\BibitemShut {NoStop}%
\bibitem [{\citenamefont {Skilling}(2006)}]{skilling2006nested}%
  \BibitemOpen
  \bibfield  {author} {\bibinfo {author} {\bibfnamefont {J.}~\bibnamefont
  {Skilling}},\ }\bibfield  {title} {\bibinfo {title} {Nested sampling for
  general bayesian computation},\ }\href@noop {} {\  (\bibinfo {year}
  {2006})}\BibitemShut {NoStop}%
\bibitem [{\citenamefont {Feroz}\ \emph {et~al.}(2009)\citenamefont {Feroz},
  \citenamefont {Hobson},\ and\ \citenamefont {Bridges}}]{feroz2009multinest}%
  \BibitemOpen
  \bibfield  {author} {\bibinfo {author} {\bibfnamefont {F.}~\bibnamefont
  {Feroz}}, \bibinfo {author} {\bibfnamefont {M.}~\bibnamefont {Hobson}},\ and\
  \bibinfo {author} {\bibfnamefont {M.}~\bibnamefont {Bridges}},\ }\bibfield
  {title} {\bibinfo {title} {Multinest: an efficient and robust bayesian
  inference tool for cosmology and particle physics},\ }\href@noop {}
  {\bibfield  {journal} {\bibinfo  {journal} {Monthly Notices of the Royal
  Astronomical Society}\ }\textbf {\bibinfo {volume} {398}},\ \bibinfo {pages}
  {1601} (\bibinfo {year} {2009})}\BibitemShut {NoStop}%
\bibitem [{\citenamefont {P{\'a}rtay}\ \emph {et~al.}(2010)\citenamefont
  {P{\'a}rtay}, \citenamefont {Bart{\'o}k},\ and\ \citenamefont
  {Cs{\'a}nyi}}]{partay2010efficient}%
  \BibitemOpen
  \bibfield  {author} {\bibinfo {author} {\bibfnamefont {L.~B.}\ \bibnamefont
  {P{\'a}rtay}}, \bibinfo {author} {\bibfnamefont {A.~P.}\ \bibnamefont
  {Bart{\'o}k}},\ and\ \bibinfo {author} {\bibfnamefont {G.}~\bibnamefont
  {Cs{\'a}nyi}},\ }\bibfield  {title} {\bibinfo {title} {Efficient sampling of
  atomic configurational spaces},\ }\href@noop {} {\bibfield  {journal}
  {\bibinfo  {journal} {The Journal of Physical Chemistry B}\ }\textbf
  {\bibinfo {volume} {114}},\ \bibinfo {pages} {10502} (\bibinfo {year}
  {2010})}\BibitemShut {NoStop}%
\bibitem [{\citenamefont {P{\'a}rtay}\ \emph {et~al.}(2021)\citenamefont
  {P{\'a}rtay}, \citenamefont {Cs{\'a}nyi},\ and\ \citenamefont
  {Bernstein}}]{partay2021nested}%
  \BibitemOpen
  \bibfield  {author} {\bibinfo {author} {\bibfnamefont {L.~B.}\ \bibnamefont
  {P{\'a}rtay}}, \bibinfo {author} {\bibfnamefont {G.}~\bibnamefont
  {Cs{\'a}nyi}},\ and\ \bibinfo {author} {\bibfnamefont {N.}~\bibnamefont
  {Bernstein}},\ }\bibfield  {title} {\bibinfo {title} {Nested sampling for
  materials},\ }\href@noop {} {\bibfield  {journal} {\bibinfo  {journal} {The
  European Physical Journal B}\ }\textbf {\bibinfo {volume} {94}},\ \bibinfo
  {pages} {159} (\bibinfo {year} {2021})}\BibitemShut {NoStop}%
\bibitem [{\citenamefont {Ashton}\ \emph {et~al.}(2022)\citenamefont {Ashton},
  \citenamefont {Bernstein}, \citenamefont {Buchner}, \citenamefont {Chen},
  \citenamefont {Cs{\'a}nyi}, \citenamefont {Fowlie}, \citenamefont {Feroz},
  \citenamefont {Griffiths}, \citenamefont {Handley}, \citenamefont {Habeck}
  \emph {et~al.}}]{ashton2022nested}%
  \BibitemOpen
  \bibfield  {author} {\bibinfo {author} {\bibfnamefont {G.}~\bibnamefont
  {Ashton}}, \bibinfo {author} {\bibfnamefont {N.}~\bibnamefont {Bernstein}},
  \bibinfo {author} {\bibfnamefont {J.}~\bibnamefont {Buchner}}, \bibinfo
  {author} {\bibfnamefont {X.}~\bibnamefont {Chen}}, \bibinfo {author}
  {\bibfnamefont {G.}~\bibnamefont {Cs{\'a}nyi}}, \bibinfo {author}
  {\bibfnamefont {A.}~\bibnamefont {Fowlie}}, \bibinfo {author} {\bibfnamefont
  {F.}~\bibnamefont {Feroz}}, \bibinfo {author} {\bibfnamefont
  {M.}~\bibnamefont {Griffiths}}, \bibinfo {author} {\bibfnamefont
  {W.}~\bibnamefont {Handley}}, \bibinfo {author} {\bibfnamefont
  {M.}~\bibnamefont {Habeck}}, \emph {et~al.},\ }\bibfield  {title} {\bibinfo
  {title} {Nested sampling for physical scientists},\ }\href@noop {} {\bibfield
   {journal} {\bibinfo  {journal} {Nature Reviews Methods Primers}\ }\textbf
  {\bibinfo {volume} {2}},\ \bibinfo {pages} {39} (\bibinfo {year}
  {2022})}\BibitemShut {NoStop}%
\bibitem [{\citenamefont {Buchner}(2023)}]{buchner2023nested}%
  \BibitemOpen
  \bibfield  {author} {\bibinfo {author} {\bibfnamefont {J.}~\bibnamefont
  {Buchner}},\ }\bibfield  {title} {\bibinfo {title} {Nested sampling
  methods},\ }\href@noop {} {\bibfield  {journal} {\bibinfo  {journal}
  {Statistic Surveys}\ }\textbf {\bibinfo {volume} {17}},\ \bibinfo {pages}
  {169} (\bibinfo {year} {2023})}\BibitemShut {NoStop}%
\bibitem [{\citenamefont {Skilling}(2012)}]{skilling2012bayesian}%
  \BibitemOpen
  \bibfield  {author} {\bibinfo {author} {\bibfnamefont {J.}~\bibnamefont
  {Skilling}},\ }\bibfield  {title} {\bibinfo {title} {Bayesian computation in
  big spaces-nested sampling and galilean monte carlo},\ }in\ \href@noop {}
  {\emph {\bibinfo {booktitle} {AIP Conference Proceedings 31st}}},\ Vol.\
  \bibinfo {volume} {1443}\ (\bibinfo {organization} {American Institute of
  Physics},\ \bibinfo {year} {2012})\ pp.\ \bibinfo {pages}
  {145--156}\BibitemShut {NoStop}%
\bibitem [{\citenamefont {Skilling}(2019)}]{skilling2019galilean}%
  \BibitemOpen
  \bibfield  {author} {\bibinfo {author} {\bibfnamefont {J.}~\bibnamefont
  {Skilling}},\ }\bibfield  {title} {\bibinfo {title} {Galilean and hamiltonian
  monte carlo},\ }in\ \href@noop {} {\emph {\bibinfo {booktitle}
  {Proceedings}}},\ Vol.~\bibinfo {volume} {33}\ (\bibinfo {organization}
  {MDPI},\ \bibinfo {year} {2019})\ p.~\bibinfo {pages} {19}\BibitemShut
  {NoStop}%
\bibitem [{\citenamefont {Lemos}\ \emph {et~al.}(2023)\citenamefont {Lemos},
  \citenamefont {Malkin}, \citenamefont {Handley}, \citenamefont {Bengio},
  \citenamefont {Hezaveh},\ and\ \citenamefont
  {Perreault-Levasseur}}]{lemos2023improving}%
  \BibitemOpen
  \bibfield  {author} {\bibinfo {author} {\bibfnamefont {P.}~\bibnamefont
  {Lemos}}, \bibinfo {author} {\bibfnamefont {N.}~\bibnamefont {Malkin}},
  \bibinfo {author} {\bibfnamefont {W.}~\bibnamefont {Handley}}, \bibinfo
  {author} {\bibfnamefont {Y.}~\bibnamefont {Bengio}}, \bibinfo {author}
  {\bibfnamefont {Y.}~\bibnamefont {Hezaveh}},\ and\ \bibinfo {author}
  {\bibfnamefont {L.}~\bibnamefont {Perreault-Levasseur}},\ }\bibfield  {title}
  {\bibinfo {title} {Improving gradient-guided nested sampling for posterior
  inference},\ }\href@noop {} {\bibfield  {journal} {\bibinfo  {journal} {arXiv
  preprint arXiv:2312.03911}\ } (\bibinfo {year} {2023})}\BibitemShut {NoStop}%
\bibitem [{\citenamefont {Olander}(2020)}]{olander2020constrained}%
  \BibitemOpen
  \bibfield  {author} {\bibinfo {author} {\bibfnamefont {J.}~\bibnamefont
  {Olander}},\ }\bibfield  {title} {\bibinfo {title} {Constrained space mcmc
  methods for nested sampling bayesian computations},\ }\href@noop {} {\
  (\bibinfo {year} {2020})}\BibitemShut {NoStop}%
\bibitem [{\citenamefont {Smith}(1984)}]{smith1984efficient}%
  \BibitemOpen
  \bibfield  {author} {\bibinfo {author} {\bibfnamefont {R.~L.}\ \bibnamefont
  {Smith}},\ }\bibfield  {title} {\bibinfo {title} {Efficient monte carlo
  procedures for generating points uniformly distributed over bounded
  regions},\ }\href@noop {} {\bibfield  {journal} {\bibinfo  {journal}
  {Operations Research}\ }\textbf {\bibinfo {volume} {32}},\ \bibinfo {pages}
  {1296} (\bibinfo {year} {1984})}\BibitemShut {NoStop}%
\bibitem [{\citenamefont {Handley}\ \emph {et~al.}(2015)\citenamefont
  {Handley}, \citenamefont {Hobson},\ and\ \citenamefont
  {Lasenby}}]{handley2015polychord}%
  \BibitemOpen
  \bibfield  {author} {\bibinfo {author} {\bibfnamefont {W.}~\bibnamefont
  {Handley}}, \bibinfo {author} {\bibfnamefont {M.}~\bibnamefont {Hobson}},\
  and\ \bibinfo {author} {\bibfnamefont {A.}~\bibnamefont {Lasenby}},\
  }\bibfield  {title} {\bibinfo {title} {Polychord: next-generation nested
  sampling},\ }\href@noop {} {\bibfield  {journal} {\bibinfo  {journal}
  {Monthly Notices of the Royal Astronomical Society}\ }\textbf {\bibinfo
  {volume} {453}},\ \bibinfo {pages} {4384} (\bibinfo {year}
  {2015})}\BibitemShut {NoStop}%
\bibitem [{\citenamefont {Baydin}\ \emph {et~al.}(2018)\citenamefont {Baydin},
  \citenamefont {Pearlmutter}, \citenamefont {Radul},\ and\ \citenamefont
  {Siskind}}]{baydin2018automatic}%
  \BibitemOpen
  \bibfield  {author} {\bibinfo {author} {\bibfnamefont {A.~G.}\ \bibnamefont
  {Baydin}}, \bibinfo {author} {\bibfnamefont {B.~A.}\ \bibnamefont
  {Pearlmutter}}, \bibinfo {author} {\bibfnamefont {A.~A.}\ \bibnamefont
  {Radul}},\ and\ \bibinfo {author} {\bibfnamefont {J.~M.}\ \bibnamefont
  {Siskind}},\ }\bibfield  {title} {\bibinfo {title} {Automatic differentiation
  in machine learning: a survey},\ }\href@noop {} {\bibfield  {journal}
  {\bibinfo  {journal} {Journal of machine learning research}\ }\textbf
  {\bibinfo {volume} {18}},\ \bibinfo {pages} {1} (\bibinfo {year}
  {2018})}\BibitemShut {NoStop}%
\bibitem [{\citenamefont {Paszke}\ \emph {et~al.}(2019)\citenamefont {Paszke},
  \citenamefont {Gross}, \citenamefont {Massa}, \citenamefont {Lerer},
  \citenamefont {Bradbury}, \citenamefont {Chanan}, \citenamefont {Killeen},
  \citenamefont {Lin}, \citenamefont {Gimelshein}, \citenamefont {Antiga} \emph
  {et~al.}}]{paszke2019pytorch}%
  \BibitemOpen
  \bibfield  {author} {\bibinfo {author} {\bibfnamefont {A.}~\bibnamefont
  {Paszke}}, \bibinfo {author} {\bibfnamefont {S.}~\bibnamefont {Gross}},
  \bibinfo {author} {\bibfnamefont {F.}~\bibnamefont {Massa}}, \bibinfo
  {author} {\bibfnamefont {A.}~\bibnamefont {Lerer}}, \bibinfo {author}
  {\bibfnamefont {J.}~\bibnamefont {Bradbury}}, \bibinfo {author}
  {\bibfnamefont {G.}~\bibnamefont {Chanan}}, \bibinfo {author} {\bibfnamefont
  {T.}~\bibnamefont {Killeen}}, \bibinfo {author} {\bibfnamefont
  {Z.}~\bibnamefont {Lin}}, \bibinfo {author} {\bibfnamefont {N.}~\bibnamefont
  {Gimelshein}}, \bibinfo {author} {\bibfnamefont {L.}~\bibnamefont {Antiga}},
  \emph {et~al.},\ }\bibfield  {title} {\bibinfo {title} {Pytorch: An
  imperative style, high-performance deep learning library},\ }\href@noop {}
  {\bibfield  {journal} {\bibinfo  {journal} {Advances in neural information
  processing systems}\ }\textbf {\bibinfo {volume} {32}} (\bibinfo {year}
  {2019})}\BibitemShut {NoStop}%
\bibitem [{\citenamefont {Ruiz-Zapatero}\ \emph {et~al.}(2023)\citenamefont
  {Ruiz-Zapatero}, \citenamefont {Hadzhiyska}, \citenamefont {Alonso},
  \citenamefont {Ferreira}, \citenamefont {Garc{\'\i}a-Garc{\'\i}a},\ and\
  \citenamefont {Mootoovaloo}}]{ruiz2023analytical}%
  \BibitemOpen
  \bibfield  {author} {\bibinfo {author} {\bibfnamefont {J.}~\bibnamefont
  {Ruiz-Zapatero}}, \bibinfo {author} {\bibfnamefont {B.}~\bibnamefont
  {Hadzhiyska}}, \bibinfo {author} {\bibfnamefont {D.}~\bibnamefont {Alonso}},
  \bibinfo {author} {\bibfnamefont {P.~G.}\ \bibnamefont {Ferreira}}, \bibinfo
  {author} {\bibfnamefont {C.}~\bibnamefont {Garc{\'\i}a-Garc{\'\i}a}},\ and\
  \bibinfo {author} {\bibfnamefont {A.}~\bibnamefont {Mootoovaloo}},\
  }\bibfield  {title} {\bibinfo {title} {Analytical marginalization over
  photometric redshift uncertainties in cosmic shear analyses},\ }\href@noop {}
  {\bibfield  {journal} {\bibinfo  {journal} {Monthly Notices of the Royal
  Astronomical Society}\ }\textbf {\bibinfo {volume} {522}},\ \bibinfo {pages}
  {5037} (\bibinfo {year} {2023})}\BibitemShut {NoStop}%
\bibitem [{\citenamefont {Unke}\ \emph {et~al.}(2021)\citenamefont {Unke},
  \citenamefont {Chmiela}, \citenamefont {Sauceda}, \citenamefont {Gastegger},
  \citenamefont {Poltavsky}, \citenamefont {Schütt}, \citenamefont
  {Tkatchenko},\ and\ \citenamefont {Müller}}]{unke2021machine}%
  \BibitemOpen
  \bibfield  {author} {\bibinfo {author} {\bibfnamefont {O.~T.}\ \bibnamefont
  {Unke}}, \bibinfo {author} {\bibfnamefont {S.}~\bibnamefont {Chmiela}},
  \bibinfo {author} {\bibfnamefont {H.~E.}\ \bibnamefont {Sauceda}}, \bibinfo
  {author} {\bibfnamefont {M.}~\bibnamefont {Gastegger}}, \bibinfo {author}
  {\bibfnamefont {I.}~\bibnamefont {Poltavsky}}, \bibinfo {author}
  {\bibfnamefont {K.~T.}\ \bibnamefont {Schütt}}, \bibinfo {author}
  {\bibfnamefont {A.}~\bibnamefont {Tkatchenko}},\ and\ \bibinfo {author}
  {\bibfnamefont {K.-R.}\ \bibnamefont {Müller}},\ }\bibfield  {title}
  {\bibinfo {title} {Machine learning force fields},\ }\href@noop {} {\bibfield
   {journal} {\bibinfo  {journal} {Chemical Reviews}\ }\textbf {\bibinfo
  {volume} {121}},\ \bibinfo {pages} {10142} (\bibinfo {year}
  {2021})}\BibitemShut {NoStop}%
\bibitem [{\citenamefont {Brooks}\ \emph {et~al.}(2011)\citenamefont {Brooks},
  \citenamefont {Gelman}, \citenamefont {Jones},\ and\ \citenamefont
  {Meng}}]{brooks2011handbook}%
  \BibitemOpen
  \bibfield  {author} {\bibinfo {author} {\bibfnamefont {S.}~\bibnamefont
  {Brooks}}, \bibinfo {author} {\bibfnamefont {A.}~\bibnamefont {Gelman}},
  \bibinfo {author} {\bibfnamefont {G.}~\bibnamefont {Jones}},\ and\ \bibinfo
  {author} {\bibfnamefont {X.-L.}\ \bibnamefont {Meng}},\ }\href@noop {} {\emph
  {\bibinfo {title} {Handbook of markov chain monte carlo}}}\ (\bibinfo
  {publisher} {CRC press},\ \bibinfo {year} {2011})\BibitemShut {NoStop}%
\bibitem [{\citenamefont {Stillinger}(1999)}]{stillinger1999exponential}%
  \BibitemOpen
  \bibfield  {author} {\bibinfo {author} {\bibfnamefont {F.~H.}\ \bibnamefont
  {Stillinger}},\ }\bibfield  {title} {\bibinfo {title} {Exponential
  multiplicity of inherent structures},\ }\href@noop {} {\bibfield  {journal}
  {\bibinfo  {journal} {Physical Review E}\ }\textbf {\bibinfo {volume} {59}},\
  \bibinfo {pages} {48} (\bibinfo {year} {1999})}\BibitemShut {NoStop}%
\bibitem [{\citenamefont {Hoare}\ and\ \citenamefont
  {McInnes}(1976)}]{hoare1976statistical}%
  \BibitemOpen
  \bibfield  {author} {\bibinfo {author} {\bibfnamefont {M.}~\bibnamefont
  {Hoare}}\ and\ \bibinfo {author} {\bibfnamefont {J.}~\bibnamefont
  {McInnes}},\ }\bibfield  {title} {\bibinfo {title} {Statistical mechanics and
  morphology of very small atomic clusters},\ }\href@noop {} {\bibfield
  {journal} {\bibinfo  {journal} {Faraday Discussions of the Chemical Society}\
  }\textbf {\bibinfo {volume} {61}},\ \bibinfo {pages} {12} (\bibinfo {year}
  {1976})}\BibitemShut {NoStop}%
\bibitem [{\citenamefont {Tsai}\ and\ \citenamefont
  {Jordan}(1993)}]{tsai1993use}%
  \BibitemOpen
  \bibfield  {author} {\bibinfo {author} {\bibfnamefont {C.}~\bibnamefont
  {Tsai}}\ and\ \bibinfo {author} {\bibfnamefont {K.}~\bibnamefont {Jordan}},\
  }\bibfield  {title} {\bibinfo {title} {Use of an eigenmode method to locate
  the stationary points on the potential energy surfaces of selected argon and
  water clusters},\ }\href@noop {} {\bibfield  {journal} {\bibinfo  {journal}
  {The Journal of Physical Chemistry}\ }\textbf {\bibinfo {volume} {97}},\
  \bibinfo {pages} {11227} (\bibinfo {year} {1993})}\BibitemShut {NoStop}%
\bibitem [{\citenamefont {Pickard}\ and\ \citenamefont
  {Needs}(2011)}]{pickard2011ab}%
  \BibitemOpen
  \bibfield  {author} {\bibinfo {author} {\bibfnamefont {C.~J.}\ \bibnamefont
  {Pickard}}\ and\ \bibinfo {author} {\bibfnamefont {R.}~\bibnamefont
  {Needs}},\ }\bibfield  {title} {\bibinfo {title} {Ab initio random structure
  searching},\ }\href@noop {} {\bibfield  {journal} {\bibinfo  {journal}
  {Journal of Physics: Condensed Matter}\ }\textbf {\bibinfo {volume} {23}},\
  \bibinfo {pages} {053201} (\bibinfo {year} {2011})}\BibitemShut {NoStop}%
\bibitem [{\citenamefont {Lov{\'a}sz}\ and\ \citenamefont
  {Vempala}(2004)}]{lovasz2004hit}%
  \BibitemOpen
  \bibfield  {author} {\bibinfo {author} {\bibfnamefont {L.}~\bibnamefont
  {Lov{\'a}sz}}\ and\ \bibinfo {author} {\bibfnamefont {S.}~\bibnamefont
  {Vempala}},\ }\bibfield  {title} {\bibinfo {title} {Hit-and-run from a
  corner},\ }in\ \href@noop {} {\emph {\bibinfo {booktitle} {Proceedings of the
  thirty-sixth annual ACM symposium on Theory of computing}}}\ (\bibinfo {year}
  {2004})\ pp.\ \bibinfo {pages} {310--314}\BibitemShut {NoStop}%
\bibitem [{\citenamefont {Lov{\'a}sz}\ and\ \citenamefont
  {Simonovits}(1993)}]{lovasz1993random}%
  \BibitemOpen
  \bibfield  {author} {\bibinfo {author} {\bibfnamefont {L.}~\bibnamefont
  {Lov{\'a}sz}}\ and\ \bibinfo {author} {\bibfnamefont {M.}~\bibnamefont
  {Simonovits}},\ }\bibfield  {title} {\bibinfo {title} {Random walks in a
  convex body and an improved volume algorithm},\ }\href@noop {} {\bibfield
  {journal} {\bibinfo  {journal} {Random structures \& algorithms}\ }\textbf
  {\bibinfo {volume} {4}},\ \bibinfo {pages} {359} (\bibinfo {year}
  {1993})}\BibitemShut {NoStop}%
\bibitem [{\citenamefont {Kannan}\ \emph {et~al.}(2006)\citenamefont {Kannan},
  \citenamefont {Lov{\'a}sz},\ and\ \citenamefont
  {Montenegro}}]{kannan2006blocking}%
  \BibitemOpen
  \bibfield  {author} {\bibinfo {author} {\bibfnamefont {R.}~\bibnamefont
  {Kannan}}, \bibinfo {author} {\bibfnamefont {L.}~\bibnamefont {Lov{\'a}sz}},\
  and\ \bibinfo {author} {\bibfnamefont {R.}~\bibnamefont {Montenegro}},\
  }\bibfield  {title} {\bibinfo {title} {Blocking conductance and mixing in
  random walks},\ }\href@noop {} {\bibfield  {journal} {\bibinfo  {journal}
  {Combinatorics, Probability and Computing}\ }\textbf {\bibinfo {volume}
  {15}},\ \bibinfo {pages} {541} (\bibinfo {year} {2006})}\BibitemShut
  {NoStop}%
\bibitem [{\citenamefont {Lov{\'a}sz}\ and\ \citenamefont
  {Simonovits}(1990)}]{lovasz1990mixing}%
  \BibitemOpen
  \bibfield  {author} {\bibinfo {author} {\bibfnamefont {L.}~\bibnamefont
  {Lov{\'a}sz}}\ and\ \bibinfo {author} {\bibfnamefont {M.}~\bibnamefont
  {Simonovits}},\ }\bibfield  {title} {\bibinfo {title} {The mixing rate of
  markov chains, an isoperimetric inequality, and computing the volume},\ }in\
  \href@noop {} {\emph {\bibinfo {booktitle} {Proceedings [1990] 31st annual
  symposium on foundations of computer science}}}\ (\bibinfo {organization}
  {IEEE},\ \bibinfo {year} {1990})\ pp.\ \bibinfo {pages}
  {346--354}\BibitemShut {NoStop}%
\bibitem [{\citenamefont {Vempala}(2005)}]{vempala2005geometric}%
  \BibitemOpen
  \bibfield  {author} {\bibinfo {author} {\bibfnamefont {S.}~\bibnamefont
  {Vempala}},\ }\bibfield  {title} {\bibinfo {title} {Geometric random walks: a
  survey},\ }\href@noop {} {\bibfield  {journal} {\bibinfo  {journal}
  {Combinatorial and computational geometry}\ }\textbf {\bibinfo {volume}
  {52}},\ \bibinfo {pages} {2} (\bibinfo {year} {2005})}\BibitemShut {NoStop}%
\bibitem [{\citenamefont {Feroz}\ and\ \citenamefont
  {Skilling}(2013)}]{feroz2013exploring}%
  \BibitemOpen
  \bibfield  {author} {\bibinfo {author} {\bibfnamefont {F.}~\bibnamefont
  {Feroz}}\ and\ \bibinfo {author} {\bibfnamefont {J.}~\bibnamefont
  {Skilling}},\ }\bibfield  {title} {\bibinfo {title} {Exploring multi-modal
  distributions with nested sampling},\ }in\ \href@noop {} {\emph {\bibinfo
  {booktitle} {AIP Conference Proceedings}}},\ Vol.\ \bibinfo {volume} {1553}\
  (\bibinfo {organization} {American Institute of Physics},\ \bibinfo {year}
  {2013})\ pp.\ \bibinfo {pages} {106--113}\BibitemShut {NoStop}%
\bibitem [{\citenamefont {Emiris}\ and\ \citenamefont
  {Fisikopoulos}(2014)}]{emiris2014efficient}%
  \BibitemOpen
  \bibfield  {author} {\bibinfo {author} {\bibfnamefont {I.~Z.}\ \bibnamefont
  {Emiris}}\ and\ \bibinfo {author} {\bibfnamefont {V.}~\bibnamefont
  {Fisikopoulos}},\ }\bibfield  {title} {\bibinfo {title} {Efficient
  random-walk methods for approximating polytope volume},\ }in\ \href@noop {}
  {\emph {\bibinfo {booktitle} {Proceedings of the thirtieth annual symposium
  on Computational geometry}}}\ (\bibinfo {year} {2014})\ pp.\ \bibinfo {pages}
  {318--327}\BibitemShut {NoStop}%
\bibitem [{\citenamefont {Chevallier}\ \emph {et~al.}(2022)\citenamefont
  {Chevallier}, \citenamefont {Cazals},\ and\ \citenamefont
  {Fearnhead}}]{chevallier2022efficient}%
  \BibitemOpen
  \bibfield  {author} {\bibinfo {author} {\bibfnamefont {A.}~\bibnamefont
  {Chevallier}}, \bibinfo {author} {\bibfnamefont {F.}~\bibnamefont {Cazals}},\
  and\ \bibinfo {author} {\bibfnamefont {P.}~\bibnamefont {Fearnhead}},\
  }\bibfield  {title} {\bibinfo {title} {Efficient computation of the the
  volume of a polytope in high-dimensions using piecewise deterministic markov
  processes},\ }in\ \href@noop {} {\emph {\bibinfo {booktitle} {International
  Conference on Artificial Intelligence and Statistics}}}\ (\bibinfo
  {organization} {PMLR},\ \bibinfo {year} {2022})\ pp.\ \bibinfo {pages}
  {10146--10160}\BibitemShut {NoStop}%
\bibitem [{\citenamefont {Mohasel~Afshar}\ and\ \citenamefont
  {Domke}(2015)}]{mohasel2015reflection}%
  \BibitemOpen
  \bibfield  {author} {\bibinfo {author} {\bibfnamefont {H.}~\bibnamefont
  {Mohasel~Afshar}}\ and\ \bibinfo {author} {\bibfnamefont {J.}~\bibnamefont
  {Domke}},\ }\bibfield  {title} {\bibinfo {title} {Reflection, refraction, and
  hamiltonian monte carlo},\ }\href@noop {} {\bibfield  {journal} {\bibinfo
  {journal} {Advances in neural information processing systems}\ }\textbf
  {\bibinfo {volume} {28}} (\bibinfo {year} {2015})}\BibitemShut {NoStop}%
\bibitem [{\citenamefont {Hee}\ \emph {et~al.}(2016)\citenamefont {Hee},
  \citenamefont {Handley}, \citenamefont {Hobson},\ and\ \citenamefont
  {Lasenby}}]{hee2016bayesian}%
  \BibitemOpen
  \bibfield  {author} {\bibinfo {author} {\bibfnamefont {S.}~\bibnamefont
  {Hee}}, \bibinfo {author} {\bibfnamefont {W.}~\bibnamefont {Handley}},
  \bibinfo {author} {\bibfnamefont {M.~P.}\ \bibnamefont {Hobson}},\ and\
  \bibinfo {author} {\bibfnamefont {A.~N.}\ \bibnamefont {Lasenby}},\
  }\bibfield  {title} {\bibinfo {title} {Bayesian model selection without
  evidences: application to the dark energy equation-of-state},\ }\href@noop {}
  {\bibfield  {journal} {\bibinfo  {journal} {Monthly Notices of the Royal
  Astronomical Society}\ }\textbf {\bibinfo {volume} {455}},\ \bibinfo {pages}
  {2461} (\bibinfo {year} {2016})}\BibitemShut {NoStop}%
\bibitem [{\citenamefont {Kroupa}\ \emph {et~al.}(2024)\citenamefont {Kroupa},
  \citenamefont {Yallup}, \citenamefont {Handley},\ and\ \citenamefont
  {Hobson}}]{kroupa2024kernel}%
  \BibitemOpen
  \bibfield  {author} {\bibinfo {author} {\bibfnamefont {N.}~\bibnamefont
  {Kroupa}}, \bibinfo {author} {\bibfnamefont {D.}~\bibnamefont {Yallup}},
  \bibinfo {author} {\bibfnamefont {W.}~\bibnamefont {Handley}},\ and\ \bibinfo
  {author} {\bibfnamefont {M.}~\bibnamefont {Hobson}},\ }\bibfield  {title}
  {\bibinfo {title} {Kernel-, mean-, and noise-marginalized gaussian processes
  for exoplanet transits and h 0 inference},\ }\href@noop {} {\bibfield
  {journal} {\bibinfo  {journal} {Monthly Notices of the Royal Astronomical
  Society}\ }\textbf {\bibinfo {volume} {528}},\ \bibinfo {pages} {1232}
  (\bibinfo {year} {2024})}\BibitemShut {NoStop}%
\bibitem [{\citenamefont {Fowlie}\ \emph {et~al.}(2021)\citenamefont {Fowlie},
  \citenamefont {Handley},\ and\ \citenamefont {Su}}]{fowlie2021nested}%
  \BibitemOpen
  \bibfield  {author} {\bibinfo {author} {\bibfnamefont {A.}~\bibnamefont
  {Fowlie}}, \bibinfo {author} {\bibfnamefont {W.}~\bibnamefont {Handley}},\
  and\ \bibinfo {author} {\bibfnamefont {L.}~\bibnamefont {Su}},\ }\bibfield
  {title} {\bibinfo {title} {Nested sampling with plateaus},\ }\href@noop {}
  {\bibfield  {journal} {\bibinfo  {journal} {Monthly Notices of the Royal
  Astronomical Society}\ }\textbf {\bibinfo {volume} {503}},\ \bibinfo {pages}
  {1199} (\bibinfo {year} {2021})}\BibitemShut {NoStop}%
\bibitem [{sup()}]{supp}%
  \BibitemOpen
  \href@noop {} {}\bibinfo {note} {See Supplemental Material at [url] for a
  proof that the stationary distribution is uniform.}\BibitemShut {Stop}%
\bibitem [{\citenamefont {Brofos}\ and\ \citenamefont
  {Lederman}(2021{\natexlab{a}})}]{brofos2021numerical}%
  \BibitemOpen
  \bibfield  {author} {\bibinfo {author} {\bibfnamefont {J.~A.}\ \bibnamefont
  {Brofos}}\ and\ \bibinfo {author} {\bibfnamefont {R.~R.}\ \bibnamefont
  {Lederman}},\ }\bibfield  {title} {\bibinfo {title} {On numerical
  considerations for riemannian manifold hamiltonian monte carlo},\ }\href@noop
  {} {\bibfield  {journal} {\bibinfo  {journal} {arXiv preprint
  arXiv:2111.09995}\ } (\bibinfo {year} {2021}{\natexlab{a}})}\BibitemShut
  {NoStop}%
\bibitem [{\citenamefont {Brofos}\ and\ \citenamefont
  {Lederman}(2021{\natexlab{b}})}]{brofos2021evaluating}%
  \BibitemOpen
  \bibfield  {author} {\bibinfo {author} {\bibfnamefont {J.}~\bibnamefont
  {Brofos}}\ and\ \bibinfo {author} {\bibfnamefont {R.~R.}\ \bibnamefont
  {Lederman}},\ }\bibfield  {title} {\bibinfo {title} {Evaluating the implicit
  midpoint integrator for riemannian hamiltonian monte carlo},\ }in\ \href@noop
  {} {\emph {\bibinfo {booktitle} {International Conference on Machine
  Learning}}}\ (\bibinfo {organization} {PMLR},\ \bibinfo {year} {2021})\ pp.\
  \bibinfo {pages} {1072--1081}\BibitemShut {NoStop}%
\bibitem [{\citenamefont {Dellago}\ \emph {et~al.}(1996)\citenamefont
  {Dellago}, \citenamefont {Posch},\ and\ \citenamefont
  {Hoover}}]{dellago1996lyapunov}%
  \BibitemOpen
  \bibfield  {author} {\bibinfo {author} {\bibfnamefont {C.}~\bibnamefont
  {Dellago}}, \bibinfo {author} {\bibfnamefont {H.~A.}\ \bibnamefont {Posch}},\
  and\ \bibinfo {author} {\bibfnamefont {W.~G.}\ \bibnamefont {Hoover}},\
  }\bibfield  {title} {\bibinfo {title} {Lyapunov instability in a system of
  hard disks in equilibrium and nonequilibrium steady states},\ }\href@noop {}
  {\bibfield  {journal} {\bibinfo  {journal} {Physical Review E}\ }\textbf
  {\bibinfo {volume} {53}},\ \bibinfo {pages} {1485} (\bibinfo {year}
  {1996})}\BibitemShut {NoStop}%
\bibitem [{\citenamefont {Cuturi}(2013)}]{cuturi2013sinkhorn}%
  \BibitemOpen
  \bibfield  {author} {\bibinfo {author} {\bibfnamefont {M.}~\bibnamefont
  {Cuturi}},\ }\bibfield  {title} {\bibinfo {title} {Sinkhorn distances:
  Lightspeed computation of optimal transport},\ }\href@noop {} {\bibfield
  {journal} {\bibinfo  {journal} {Advances in neural information processing
  systems}\ }\textbf {\bibinfo {volume} {26}} (\bibinfo {year}
  {2013})}\BibitemShut {NoStop}%
\bibitem [{\citenamefont {Genevay}\ \emph {et~al.}(2016)\citenamefont
  {Genevay}, \citenamefont {Cuturi}, \citenamefont {Peyr{\'e}},\ and\
  \citenamefont {Bach}}]{genevay2016stochastic}%
  \BibitemOpen
  \bibfield  {author} {\bibinfo {author} {\bibfnamefont {A.}~\bibnamefont
  {Genevay}}, \bibinfo {author} {\bibfnamefont {M.}~\bibnamefont {Cuturi}},
  \bibinfo {author} {\bibfnamefont {G.}~\bibnamefont {Peyr{\'e}}},\ and\
  \bibinfo {author} {\bibfnamefont {F.}~\bibnamefont {Bach}},\ }\bibfield
  {title} {\bibinfo {title} {Stochastic optimization for large-scale optimal
  transport},\ }\href@noop {} {\bibfield  {journal} {\bibinfo  {journal}
  {Advances in neural information processing systems}\ }\textbf {\bibinfo
  {volume} {29}} (\bibinfo {year} {2016})}\BibitemShut {NoStop}%
\bibitem [{\citenamefont {Genevay}\ \emph {et~al.}(2018)\citenamefont
  {Genevay}, \citenamefont {Peyr{\'e}},\ and\ \citenamefont
  {Cuturi}}]{genevay2018learning}%
  \BibitemOpen
  \bibfield  {author} {\bibinfo {author} {\bibfnamefont {A.}~\bibnamefont
  {Genevay}}, \bibinfo {author} {\bibfnamefont {G.}~\bibnamefont {Peyr{\'e}}},\
  and\ \bibinfo {author} {\bibfnamefont {M.}~\bibnamefont {Cuturi}},\
  }\bibfield  {title} {\bibinfo {title} {Learning generative models with
  sinkhorn divergences},\ }in\ \href@noop {} {\emph {\bibinfo {booktitle}
  {International Conference on Artificial Intelligence and Statistics}}}\
  (\bibinfo {organization} {PMLR},\ \bibinfo {year} {2018})\ pp.\ \bibinfo
  {pages} {1608--1617}\BibitemShut {NoStop}%
\bibitem [{\citenamefont {Feydy}\ \emph {et~al.}(2019)\citenamefont {Feydy},
  \citenamefont {S{\'e}journ{\'e}}, \citenamefont {Vialard}, \citenamefont
  {Amari}, \citenamefont {Trouv{\'e}},\ and\ \citenamefont
  {Peyr{\'e}}}]{feydy2019interpolating}%
  \BibitemOpen
  \bibfield  {author} {\bibinfo {author} {\bibfnamefont {J.}~\bibnamefont
  {Feydy}}, \bibinfo {author} {\bibfnamefont {T.}~\bibnamefont
  {S{\'e}journ{\'e}}}, \bibinfo {author} {\bibfnamefont {F.-X.}\ \bibnamefont
  {Vialard}}, \bibinfo {author} {\bibfnamefont {S.-i.}\ \bibnamefont {Amari}},
  \bibinfo {author} {\bibfnamefont {A.}~\bibnamefont {Trouv{\'e}}},\ and\
  \bibinfo {author} {\bibfnamefont {G.}~\bibnamefont {Peyr{\'e}}},\ }\bibfield
  {title} {\bibinfo {title} {Interpolating between optimal transport and mmd
  using sinkhorn divergences},\ }in\ \href@noop {} {\emph {\bibinfo {booktitle}
  {The 22nd International Conference on Artificial Intelligence and
  Statistics}}}\ (\bibinfo {organization} {PMLR},\ \bibinfo {year} {2019})\
  pp.\ \bibinfo {pages} {2681--2690}\BibitemShut {NoStop}%
\bibitem [{\citenamefont {Genevay}\ \emph {et~al.}(2019)\citenamefont
  {Genevay}, \citenamefont {Chizat}, \citenamefont {Bach}, \citenamefont
  {Cuturi},\ and\ \citenamefont {Peyr{\'e}}}]{genevay2019sample}%
  \BibitemOpen
  \bibfield  {author} {\bibinfo {author} {\bibfnamefont {A.}~\bibnamefont
  {Genevay}}, \bibinfo {author} {\bibfnamefont {L.}~\bibnamefont {Chizat}},
  \bibinfo {author} {\bibfnamefont {F.}~\bibnamefont {Bach}}, \bibinfo {author}
  {\bibfnamefont {M.}~\bibnamefont {Cuturi}},\ and\ \bibinfo {author}
  {\bibfnamefont {G.}~\bibnamefont {Peyr{\'e}}},\ }\bibfield  {title} {\bibinfo
  {title} {Sample complexity of sinkhorn divergences},\ }in\ \href@noop {}
  {\emph {\bibinfo {booktitle} {The 22nd international conference on artificial
  intelligence and statistics}}}\ (\bibinfo {organization} {PMLR},\ \bibinfo
  {year} {2019})\ pp.\ \bibinfo {pages} {1574--1583}\BibitemShut {NoStop}%
\bibitem [{\citenamefont {Cuturi}\ \emph {et~al.}(2022)\citenamefont {Cuturi},
  \citenamefont {Meng-Papaxanthos}, \citenamefont {Tian}, \citenamefont
  {Bunne}, \citenamefont {Davis},\ and\ \citenamefont
  {Teboul}}]{cuturi2022optimal}%
  \BibitemOpen
  \bibfield  {author} {\bibinfo {author} {\bibfnamefont {M.}~\bibnamefont
  {Cuturi}}, \bibinfo {author} {\bibfnamefont {L.}~\bibnamefont
  {Meng-Papaxanthos}}, \bibinfo {author} {\bibfnamefont {Y.}~\bibnamefont
  {Tian}}, \bibinfo {author} {\bibfnamefont {C.}~\bibnamefont {Bunne}},
  \bibinfo {author} {\bibfnamefont {G.}~\bibnamefont {Davis}},\ and\ \bibinfo
  {author} {\bibfnamefont {O.}~\bibnamefont {Teboul}},\ }\bibfield  {title}
  {\bibinfo {title} {Optimal transport tools (ott): A jax toolbox for all
  things wasserstein},\ }\href@noop {} {\bibfield  {journal} {\bibinfo
  {journal} {arXiv preprint arXiv:2201.12324}\ } (\bibinfo {year}
  {2022})}\BibitemShut {NoStop}%
\bibitem [{\citenamefont {Vershynin}(2018)}]{vershynin2018high}%
  \BibitemOpen
  \bibfield  {author} {\bibinfo {author} {\bibfnamefont {R.}~\bibnamefont
  {Vershynin}},\ }\href@noop {} {\emph {\bibinfo {title} {High-dimensional
  probability: An introduction with applications in data science}}},\
  Vol.~\bibinfo {volume} {47}\ (\bibinfo  {publisher} {Cambridge university
  press},\ \bibinfo {year} {2018})\BibitemShut {NoStop}%
\bibitem [{\citenamefont {Ott}(2002)}]{ott2002chaos}%
  \BibitemOpen
  \bibfield  {author} {\bibinfo {author} {\bibfnamefont {E.}~\bibnamefont
  {Ott}},\ }\href@noop {} {\emph {\bibinfo {title} {Chaos in dynamical
  systems}}}\ (\bibinfo  {publisher} {Cambridge university press},\ \bibinfo
  {year} {2002})\BibitemShut {NoStop}%
\bibitem [{\citenamefont {Ball}\ \emph {et~al.}(1997)\citenamefont {Ball} \emph
  {et~al.}}]{ball1997elementary}%
  \BibitemOpen
  \bibfield  {author} {\bibinfo {author} {\bibfnamefont {K.}~\bibnamefont
  {Ball}} \emph {et~al.},\ }\bibfield  {title} {\bibinfo {title} {An elementary
  introduction to modern convex geometry},\ }\href@noop {} {\bibfield
  {journal} {\bibinfo  {journal} {Flavors of geometry}\ }\textbf {\bibinfo
  {volume} {31}},\ \bibinfo {pages} {26} (\bibinfo {year} {1997})}\BibitemShut
  {NoStop}%
\bibitem [{\citenamefont {Cover}(1999)}]{cover1999elements}%
  \BibitemOpen
  \bibfield  {author} {\bibinfo {author} {\bibfnamefont {T.~M.}\ \bibnamefont
  {Cover}},\ }\href@noop {} {\emph {\bibinfo {title} {Elements of information
  theory}}}\ (\bibinfo  {publisher} {John Wiley \& Sons},\ \bibinfo {year}
  {1999})\BibitemShut {NoStop}%
\bibitem [{\citenamefont {Lov{\'a}sz}(1999)}]{lovasz1999hit}%
  \BibitemOpen
  \bibfield  {author} {\bibinfo {author} {\bibfnamefont {L.}~\bibnamefont
  {Lov{\'a}sz}},\ }\bibfield  {title} {\bibinfo {title} {Hit-and-run mixes
  fast},\ }\href@noop {} {\bibfield  {journal} {\bibinfo  {journal}
  {Mathematical programming}\ }\textbf {\bibinfo {volume} {86}},\ \bibinfo
  {pages} {443} (\bibinfo {year} {1999})}\BibitemShut {NoStop}%
\bibitem [{\citenamefont {Weinberger}(1956)}]{weinberger1956isoperimetric}%
  \BibitemOpen
  \bibfield  {author} {\bibinfo {author} {\bibfnamefont {H.~F.}\ \bibnamefont
  {Weinberger}},\ }\bibfield  {title} {\bibinfo {title} {An isoperimetric
  inequality for the n-dimensional free membrane problem},\ }\href@noop {}
  {\bibfield  {journal} {\bibinfo  {journal} {Journal of Rational Mechanics and
  Analysis}\ }\textbf {\bibinfo {volume} {5}},\ \bibinfo {pages} {633}
  (\bibinfo {year} {1956})}\BibitemShut {NoStop}%
\bibitem [{\citenamefont {M{\"u}ller}(2012)}]{muller2012analysis}%
  \BibitemOpen
  \bibfield  {author} {\bibinfo {author} {\bibfnamefont {C.}~\bibnamefont
  {M{\"u}ller}},\ }\href@noop {} {\emph {\bibinfo {title} {Analysis of
  spherical symmetries in Euclidean spaces}}},\ Vol.\ \bibinfo {volume} {129}\
  (\bibinfo  {publisher} {Springer Science \& Business Media},\ \bibinfo {year}
  {2012})\BibitemShut {NoStop}%
\bibitem [{\citenamefont {Chen}\ and\ \citenamefont
  {Eldan}(2022)}]{chen2022hit}%
  \BibitemOpen
  \bibfield  {author} {\bibinfo {author} {\bibfnamefont {Y.}~\bibnamefont
  {Chen}}\ and\ \bibinfo {author} {\bibfnamefont {R.}~\bibnamefont {Eldan}},\
  }\bibfield  {title} {\bibinfo {title} {Hit-and-run mixing via localization
  schemes},\ }\href@noop {} {\bibfield  {journal} {\bibinfo  {journal} {arXiv
  preprint arXiv:2212.00297}\ } (\bibinfo {year} {2022})}\BibitemShut {NoStop}%
\end{thebibliography}%

\clearpage
\foreach \x in {1,2,3}
{%
	\clearpage
	

\section*{Supplementary material}

# Resonances in reflective Hamiltonian Monte Carlo

Supplemental Material

We ensure mathematical correctness of the algorithm, i.e. if the Markov chain defined by Galilean Monte Carlo (GMC) has the desired stationary distribution. We start by showing that any reflective sampler with inexact reflections has the desired stationary distribution and specialize this to the case of GMC afterwards. The desired stationary distribution $\pi$ is the uniform distribution on the volume $\mathcal{U} \subset \mathbb{R}^n$ and is written as a product distribution on coordinate and momentum space as

$$\pi(\mathbf{q}, \mathbf{p}) = \frac{1}{N} \mathbb{1}_\mathcal{U}(\mathbf{q}) \exp(-K(\mathbf{p})), \tag{1}$$

where $N$ is a normalisation constant, $\mathbb{1}_\mathcal{U}$ is the indicator function on $\mathcal{U}$ and $K(\mathbf{p}) = \frac{1}{2}|\mathbf{p}|^2 = \frac{1}{2}\sum_{i=1}^n p_i^2$ is the kinetic energy.

The requirements for correctness are formulated in terms of the flow map $\mathbf{\Phi}_t : (\mathbf{q}_0, \mathbf{p}_0) \mapsto (\mathbf{q}_t, \mathbf{p}_t)$ which deterministically maps an initial point $(\mathbf{q}_0, \mathbf{p}_0)$ in phase space to the point $(\mathbf{q}_t, \mathbf{p}_t)$ at time $t$. The two technical requirements [1] are then that the flow map is self-inverse, $\mathbf{\Phi}_t(\mathbf{\Phi}_t(\mathbf{q}, \mathbf{p})) = (\mathbf{q}, \mathbf{p})$, and has unit Jacobian, $|\det \mathbf{\Phi}'(\mathbf{q}, \mathbf{p})| = 1$, for all $(\mathbf{q}, \mathbf{p})$. The existence of the Jacobian additionally requires sufficient smoothness of $\mathbf{\Phi}$. For the example of GMC, the flow map consists of the chain of $L$ steps and an additional momentum reversal, $\mathbf{p} \mapsto -\mathbf{p}$. As in Hamiltonian Monte Carlo, the final momentum reversal is neglected in practice as $\mathbf{p}$ is re-randomized immediately afterwards [2].

In the following, the set $\mathcal{V} = \{(\mathbf{q}, \mathbf{p}) \mid \mathbf{q} \in \mathcal{U} \text{ and } \mathbf{p} \in \mathbb{R}^n\}$ denotes the region in phase space for which $\mathbf{q}$ is in $\mathcal{U}$. Furthermore, the Markov Chain defined by the reflective samplers can be written in terms of a transition kernel $T$, where $B$ is a Borel set in $\mathbb{R}^{2n}$ and $\mathbb{1}[S]$ is 1 if the statement $S$ is true and 0 otherwise,

$$T((\mathbf{q}, \mathbf{p}), B) = \underbrace{\mathbb{1}[\mathbf{\Phi}(\mathbf{q}, \mathbf{p}) \in \mathcal{V}]}_{\text{Metropolis accept}} \mathbb{1}[\mathbf{\Phi}(\mathbf{q}, \mathbf{p}) \in B] + \underbrace{\mathbb{1}[\mathbf{\Phi}(\mathbf{q}, \mathbf{p}) \notin \mathcal{V}]}_{\text{Metropolis reject}} \mathbb{1}[(\mathbf{q}, \mathbf{p}) \in B], \tag{2}$$

which states that, given an initial point $(\mathbf{q}, \mathbf{p})$, the final point is in $B$ if $\mathbf{\Phi}$ maps the initial point into $\mathcal{V}$, in which case the proposal is accepted and $\mathbf{\Phi}(\mathbf{q}, \mathbf{p})$ must lie in $B$, or if $\mathbf{\Phi}$ maps the initial point outside $\mathcal{V}$, in which case the point is rejected and the initial point must lie in $B$.

The following Theorem states that reflection-based samplers indeed sample the uniform distribution on $\mathcal{U}$.

**Theorem.** *Let $\mathbf{\Phi}$ be a self-inverse and volume-preserving map on $\mathbf{z} = (\mathbf{p}, \mathbf{q}) \in \mathbb{R}^{2n}$. Further, let the kinetic energy $K$ be invariant under $\mathbf{\Phi}$. Consider the probability distribution $\pi$ defined in Equation 1 and the Markov transition kernel $T$ defined in Equation 2. Then the stationary distribution of $T$ is $\pi$.*

*Proof.* This proof follows the approach in [1] and [3]. Let $\mathbf{z} \sim \pi(\cdot)$ and $\mathbf{z}' \sim T(\mathbf{z}, \cdot)$. Further, let $A$ and $B$ be Borel subsets of $\mathbb{R}^{2n}$. Using the definition of $T$,

$$P(\mathbf{z} \in A \text{ and } \mathbf{z}' \in B) = \int_A \pi(\mathbf{z}) T(\mathbf{z}, B) d\mathbf{z} \tag{3}$$

$$= \int_A \pi(\mathbf{z}) \mathbb{1}[\mathbf{\Phi}(\mathbf{z}) \in \mathcal{V}] \mathbb{1}[\mathbf{\Phi}(\mathbf{z}) \in B] d\mathbf{z} + \int_A \pi(\mathbf{z}) \mathbb{1}[\mathbf{\Phi}(\mathbf{z}) \notin \mathcal{V}] \mathbb{1}[\mathbf{z} \in B] d\mathbf{z}. \tag{4}$$

1

The first term becomes

$$\int_A \pi(\mathbf{z})\mathbb{1}[\mathbf{\Phi}(\mathbf{z}) \in \mathcal{V}]\mathbb{1}[\mathbf{\Phi}(\mathbf{z}) \in B]d\mathbf{z} = \int_{A \cap \mathbf{\Phi}(B)} \pi(\mathbf{z})\mathbb{1}[\mathbf{\Phi}(\mathbf{z}) \in \mathcal{V}]d\mathbf{z} \quad (5)$$

$$= \int_{\mathbf{\Phi}(A) \cap B} \pi(\mathbf{\Phi}(\mathbf{z}'))\mathbb{1}[\mathbf{z}' \in \mathcal{V}]d\mathbf{z}' \quad (6)$$

$$= \int_{\mathbf{\Phi}(A) \cap B} \pi(\mathbf{z}')\mathbb{1}[\mathbf{\Phi}(\mathbf{z}') \in \mathcal{V}]d\mathbf{z}' \quad (7)$$

$$= \int_B \pi(\mathbf{z}')\mathbb{1}[\mathbf{\Phi}(\mathbf{z}') \in \mathcal{V}]\mathbb{1}[\mathbf{z}' \in \mathbf{\Phi}(A)]d\mathbf{z}', \quad (8)$$

where $\mathbf{\Phi}(B)$ is the image of $B$, $\mathbf{\Phi} = \mathbf{\Phi}^{-1}$ was used, a change of variables with unit Jacobian to $\mathbf{z}' = \mathbf{\Phi}(\mathbf{z})$ was performed and the relation

$$\pi(\mathbf{\Phi}(\mathbf{z}'))\mathbb{1}[\mathbf{z}' \in \mathcal{V}] = \frac{1}{N}\mathbb{1}[\mathbf{\Phi}_q(\mathbf{z}') \in \mathcal{U}]\exp(-K(\mathbf{\Phi}_p(\mathbf{z}')))\mathbb{1}[\mathbf{z}' \in \mathcal{V}] \quad (9)$$

$$= \pi(\mathbf{z}')\mathbb{1}[\mathbf{\Phi}(\mathbf{z}') \in \mathcal{V}] \quad (10)$$

holds since $K(\mathbf{\Phi}_p(\mathbf{z})) = K(\mathbf{p})$ by assumption of the invariance of $K$ under $\mathbf{\Phi}$, where $\mathbf{\Phi}_q(\mathbf{z})$ and $\mathbf{\Phi}_p(\mathbf{z})$ are the position and momentum part of $\mathbf{\Phi}(\mathbf{z})$, respectively. The second term becomes

$$\int_A \pi(\mathbf{z})\mathbb{1}[\mathbf{\Phi}(\mathbf{z}) \notin \mathcal{V}]\mathbb{1}[\mathbf{z} \in B]d\mathbf{z} = \int_B \pi(\mathbf{z})\mathbb{1}[\mathbf{\Phi}(\mathbf{z}) \notin \mathcal{V}]\mathbb{1}[\mathbf{z} \in A]d\mathbf{z}. \quad (11)$$

Adding the terms back together and using the definition of $T$ gives

$$P(\mathbf{z} \in A \text{ and } \mathbf{z}' \in B) = P(\mathbf{z} \in B \text{ and } \mathbf{z}' \in A). \quad (12)$$

Setting $A = \mathbb{R}^{2d}$, this reduces to

$$P(\mathbf{z}' \in B) = P(\mathbf{z} \in B). \quad (13)$$

In summary, if $\mathbf{z}$ is sampled from $\pi$ and $\mathbf{z}'$ is sampled from the Markov kernel, then $\mathbf{z}'$ is distributed as $\pi$ as well. Hence, $\pi$ is stationary under the Markov Chain. □

In GMC and its variation described in [5], the flow map always proposes an accepted point so that the Markov kernel (Equation 2) simplifies to $T((\mathbf{q}, \mathbf{p}), B) = \mathbb{1}[\mathbf{\Phi}(\mathbf{q}, \mathbf{p}) \in \mathcal{V}]\mathbb{1}[\mathbf{\Phi}(\mathbf{q}, \mathbf{p}) \in B]$, which would allow for a simplified proof. However, for the variation named reflective slice sampling, described in [6], proposed points may lead to rejections.

The above proof shows that the flow map can be unrelated to the momentum distribution, which is proportional to $\exp(-K(\mathbf{p}))$, as long as the kinetic energy is invariant under the flow map, i.e. $K(\mathbf{\Phi}_p(\mathbf{q}, \mathbf{p})) = K(\mathbf{p})$. This is a consequence of the fact that the stationary distribution is uniform and stands in contrast to Hamiltonian Monte Carlo in which the flow map is determined by the kinetic energy. We may hence choose a different momentum distribution to mitigate the curse of dimensionality. In fact, for a mass matrix $M$, the most general kinetic energy [4] invariant under the Hamiltonian flow takes the form

$$K(\mathbf{q}, \mathbf{p}) = \kappa_1\left(\mathbf{p}^\top \mathbf{M}^{-1}(\mathbf{q})\mathbf{p}\right) + \kappa_2(\mathbf{q}), \quad (14)$$

where $\kappa_1$ and $\kappa_2$ are arbitrary functions and $\mathbf{M}$ is a real symmetric matrix, provided that the reflection is changed to $\mathbf{p} \leftarrow \mathbf{p} - 2\frac{\mathbf{p}^\top \mathbf{M}^{-1}(\mathbf{q})\mathbf{n}}{\mathbf{n}\mathbf{M}^{-1}(\mathbf{q})\mathbf{n}}\mathbf{n}$.

It remains to show that GMC indeed satisfies the assumptions of the above Theorem. Reversibility for a single GMC step can be shown by considering the flow map. Now, assuming that a single step is reversible, $(R \circ \varphi) \circ (R \circ \varphi) = \text{Id}$, where $R$ is the momentum reflection operator with the property $R \circ R = \text{Id}$, $\varphi$ is the proposal of a single GMC step, Id is the identity map and $\circ$ denotes function concatenation, the self-inverse

property of a GMC chain of $L$ steps, whose proposal $\varphi^{\circ L}$ is the $L$-fold concatenation of $\varphi$, can be shown by induction:

$$(R \circ \varphi^{\circ(L+1)}) \circ (R \circ \varphi^{\circ(L+1)}) = R \circ \varphi \circ \underbrace{R \circ R}_{= \text{Id}} \circ \varphi^{\circ L} \circ R \circ \varphi^{\circ(L+1)} \quad (15)$$

$$= R \circ \varphi \circ R \circ \underbrace{(R \circ \varphi^{\circ L} \circ R \circ \varphi^{\circ L})}_{= \text{Id, by induction}} \circ \varphi \quad (16)$$

$$= R \circ \varphi \circ R \circ \varphi \quad (17)$$

$$= \text{Id}. \quad (18)$$

With the self-inverse property of the GMC flow map established, the volume-preserving property and invariance of the kinetic energy must be shown. Volume-preservation follows from the fact that motion in a straight line, as well as a reflection, have a Jacobian with an absolute determinant of unity. Invariance of the kinetic energy holds because the magnitude of the momentum is unchanged during any GMC step.

In conclusion, GMC has a uniform stationary distribution on $\mathcal{U}$, as required.

# References

[1] James A Brofos and Roy R Lederman. "On numerical considerations for riemannian manifold hamiltonian monte carlo". In: *arXiv preprint arXiv:2111.09995* (2021).

[2] Steve Brooks et al. *Handbook of markov chain monte carlo*. CRC press, 2011.

[3] James Brofos and Roy R Lederman. "Evaluating the implicit midpoint integrator for Riemannian Hamiltonian Monte Carlo". In: *International Conference on Machine Learning*. PMLR. 2021, pp. 1072–1081.

[4] Michael Betancourt and Leo C Stein. "The geometry of hamiltonian monte carlo". In: *arXiv preprint arXiv:1112.4118* (2011).

[5] John Skilling. "Galilean and Hamiltonian Monte Carlo". In: *Proceedings*. Vol. 33. 1. MDPI. 2019, p. 19.

[6] Radford M Neal. "Slice sampling". In: *The annals of statistics* 31.3 (2003), pp. 705–767.


}

\end{document}